\pdfoutput=1
\def\CAMERAREADY{}
\documentclass[11pt]{article}
\newif\ifcameraready
\ifdefined\CAMERAREADY\camerareadytrue\else\camerareadyfalse\fi %
\ifcameraready\usepackage[final]{acl}\else\usepackage[review]{acl}\fi
\usepackage{times}
\usepackage{latexsym}
\usepackage[T1]{fontenc}
\usepackage[utf8]{inputenc}
\usepackage{microtype}
\usepackage{inconsolata}
\usepackage{booktabs}
\usepackage{amsmath,amssymb,amsthm}
\usepackage{xcolor}
\usepackage{graphicx}
\usepackage{fontawesome5}
\usepackage{algorithm}
\usepackage{algpseudocode}

\newtheorem{theorem}{Theorem}
\newtheorem{definition}{Definition}
\newtheorem{proposition}{Proposition}

\newcommand{\fld}{FLD$\times$2}
\newcommand{\proofplant}{\textsc{ProofPlant}}
\ifcameraready
\newcommand{\codeurl}{https://github.com/sileod/semforge}
\else
\newcommand{\codeurl}{http://redacted}
\fi
\newcommand{\codelink}{\href{\codeurl}{\small\faGithub\,\nolinkurl{\codeurl}}}

\title{Certify What You Plant:\\Shortcut-Free Proof Depth in Logical Reasoning Data}

\ifcameraready
\author{Baptiste Juran \qquad Dimitri Kachler \qquad Damien Sileo \\
  Univ.\ Lille, Inria, CNRS, Centrale Lille, UMR 9189 CRIStAL, F-59000 Lille, France \\
  \texttt{damien.sileo@inria.fr}}
\else
\author{Anonymous}
\fi

\begin{document}
\maketitle

\begin{abstract}
Procedurally generated logic problems are used to both evaluate and train reasoning in language
models. Their difficulty is usually set by planting a proof, that is, by sampling a proof of the desired
depth and stating its premises as the problem. However, a planted proof is only one possible proof. A
problem can also have shorter proofs, or several unrelated ones, and its stated depth then overstates what
solving it requires. We measure what a problem actually requires, relative to a stated set of inference
rules. For each problem, we compute the steps every proof must make, the shortest proof, the number of
minimal proofs and the indispensable premises, and export certificates that an independent checker
verifies. Audits of eight datasets find depth that is forced (ProofWriter,
Blocksworld), problems labelled depth~5 that require about two specific rule applications (\fld{}), and answers that need no particular step
(ZebraLogic). We then construct problems whose every proof has a requested structure, using a catalog of
shortcuts proved once per rule set. Every constructed problem is checked in full and comes with its own
certificate. In training, the best data mixes depths and problem types. At equal size, such mixtures of
certified problem types, including model-checking problems that forward generation does not produce,
transfer better than \fld{}-style derivation data for Qwen3 models from 0.6B to 8B.
Code is available at \codelink.
\end{abstract}

\section{Introduction}

Deductive reasoning is one of the few abilities of language models that can be tested and taught with
exact supervision. A generator can produce logic problems in unlimited number, each with a label verified
by a solver, and such data is now used both as benchmark and as training material
\citep{clark2020ruletaker,tafjord2021proofwriter,morishita2023fld,morishita2024fldx2,lin2025zebralogic,saparov2023prontoqaood,liu2025synlogic,stojanovski2025reasoninggym}.

As models improve, shallow problems lose their value for both uses. Benchmarks saturate, and a model
learns little from problems it already solves. Generators therefore expose a \emph{depth} parameter, the
number of inference steps a solution takes, and set it by \emph{planting}. They sample a proof tree of the
desired depth, state its leaves as premises, add distractors, and report the planted depth as the
difficulty of the problem.

\begin{figure}[t]
\centering
\includegraphics[width=\columnwidth]{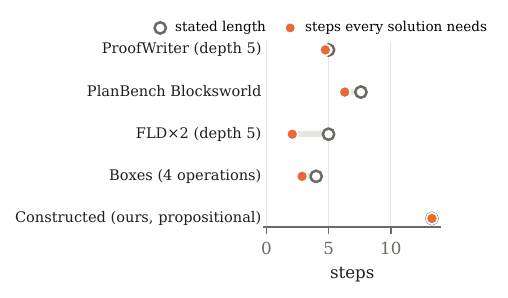}
\caption{What datasets state and what their problems require. Open circles: the stated length (planted
depth, plan length, number of operations); dots: the steps every solution needs (mandatory proof steps;
for plans, landmark actions; for state tracking, operations without which the answer changes). Bottom row:
400 problems built by our construction (\S\ref{sec:construction}) in propositional logic, with 13.3
planted steps on average; circle and dot coincide because every planted step is required
(Table~\ref{tab:proof-audit}).}
\label{fig:teaser}
\end{figure}

A planted proof shows that a problem can be solved in five steps, but not that five steps are needed. A rule that chains two implications (from $A\to B$ and $B\to C$,
infer $A\to C$) replaces two modus ponens steps, a rule that chains two general statements (all A are B,
all B are C, so all A are C) skips an intermediate conclusion about an individual, and distractors can
open routes of their own. A problem labelled ``5 steps'' may then be provable in two, or in several ways
that share no step, so that none of its five steps is required (Figure~\ref{fig:example}). We distinguish
the \emph{planted} structure of a problem, which describes one proof, from its \emph{realized} structure,
which describes what all proofs have in common. Only the second says what a solver must do, and nothing in
forward generation makes the two agree. Measured this way, several datasets ask for much less than they
state (Figure~\ref{fig:teaser}).

This leaves two questions. How deep are reasoning datasets, measured by what their problems require? And
once depth can be guaranteed, how should training data use it?

For the first question, we make realized structure a measured and certified quantity, and then a
construction target. The framework is general. A logic is declared by a signature, sound rule schemas and
a semantics decided by a solver, and the measures, the construction and the checker run on any such
declaration without logic-specific code. We instantiate fifteen, from propositional, first-order and modal
logic to constraint puzzles, planning and state tracking. Because structure is counted in the declared
rules, a reader who prefers other rules can declare them and rerun every audit. Adding rules can only make
steps avoidable, never required (Proposition~\ref{prop:mono}).

The second question turns out to be one of distribution design. At a
fixed planted depth, examples that require their steps train much like examples that do not, once a label
shortcut that comes with deep proofs is balanced. Deep-only data makes models abstain, and short proofs
keep this in check. The best training sets combine depths and kinds of problems, and their advantage over
uncertified data comes mainly from that combination. Our contributions are:
\begin{itemize}
\item \textbf{Measuring and auditing realized structure} (\S\ref{sec:measures}, \S\ref{sec:audit}).
  Relative to a stated calculus, we certify the mandatory steps of a problem (the rule applications
  present in every proof) and their topology, the shallowest and smallest proof and the number of minimal
  proofs, and, independently of any calculus, its minimal premise sets. Applied to ProofWriter, \fld{},
  ZebraLogic, corrected FOLIO, PlanBench and the Boxes state-tracking data, and at the premise level to ProverQA and LogicGraph, from
  their own logical forms, these measures separate datasets whose claimed structure is forced from
  datasets where it is optional, and name four distinct ways in which a problem can have no mandatory
  step.
\item \textbf{Constructing certified structure} (\S\ref{sec:construction}, \S\ref{sec:certificates}).
  Target-conditioned construction (\proofplant{}) builds problems whose every planned step is mandatory, at scale. It
  relies on a \emph{shortcut atlas}, a set of \emph{nogoods}\footnote{The term comes from constraint solving, where a nogood is a combination
of choices known to admit no solution. Here it is a proof fragment known to have a shortcut, so that no
problem containing it can have all of its steps mandatory.} over proof fragments proved once per calculus, with
  a soundness theorem for their use. Fifteen rule sets (propositional, first-order, modal, constraints,
  planning, state tracking) have machine-checked rules. Every generated problem comes with a certificate checked by code
  that shares nothing with the generator, and neutral labels are certified by finite countermodels.
\item \textbf{A training study of depth and distribution shape} (\S\ref{sec:transfer}). Training sets
  that differ only in certified structure show that, at a fixed planted depth, the number of mandatory steps
  alone changes little once a label shortcut it brings along is controlled, that short proofs
  are needed beside deep ones, and that the number of proofs matters little. Certified problem types also
  give the training mixture coordinates: at equal size, mixtures over them transfer better to unseen
  benchmarks than \fld{}-style data over the same logics, for Qwen3 from 0.6B to 8B, through their
  composition, including Kripke examples the baseline lacks.
\end{itemize}
Every audit is reproduced from public data by one script, and the code is available (\codelink).

\section{Related Work}

\paragraph{Synthetic reasoning corpora.}
RuleTaker and ProofWriter \citep{clark2020ruletaker,tafjord2021proofwriter} control proof depth; FLD and
\fld{} \citep{morishita2023fld,morishita2024fldx2} design a corpus by principles (multi-step deduction,
distractors, diverse rules); PrOntoQA-OOD \citep{saparov2023prontoqaood} controls rules and proof
composition, MathGAP \citep{opedal2025mathgap} proof-tree shape in arithmetic, and ScaleLogic
\citep{scalelogic2026} depth apart from expressiveness. SynLogic and Reasoning Gym
\citep{liu2025synlogic,stojanovski2025reasoninggym} scale up verifiable task families; PuzzleClone
\citep{xiong2025puzzleclone} verifies answers with an SMT solver; FOL-NLI \citep{sileo2024unigram} binds
English to a prover's language by grammars; ProverGen \citep{qi2025provergen} verifies problems and
proofs with a prover; RSS \citep{bu2025rss} samples reasoning topologies with a GFlowNet. A verified label
and a valid proof do not make that proof's steps necessary (\S\ref{sec:audit}): generators that plant a
proof do not check which of its steps are required, and grammar sampling as in FOL-NLI plants no proof to
check. We certify which planted structure is realized, and construct data so that all of it is.

\paragraph{Claimed against realized depth.}
HopToM \citep{you2026hoptom} makes the same observation for theory of mind: nested-belief questions in
existing benchmarks rarely require their nominal order, measured by the distinct events that support each
belief layer, and it releases questions certified at full order. That measure reads one deterministic
resolution trace of belief updates. Ours is relative to a stated calculus, counts alternative proofs
(mandatory steps, re-routing, over-determination), covers deduction, constraint satisfaction and
planning, is checked independently, and serves as a construction target and a training-mixture coordinate.

\paragraph{Multiple proofs.}
LogicGraph \citep{wu2026logicgraph} enumerates every minimal proof to build a benchmark of multi-path
problems (audited in \S\ref{sec:audit}). Synthology \citep{vanschependom2026synthology} builds proofs backward from ontologies. LogicSkills
\citep{rabern2026logicskills} and ProofGrid \citep{arkoudas2025proofgrid,arkoudas2026proofgrid} evaluate countermodel
construction and proof writing. We use alternative proofs as a quantity to audit and as a construction
target (exactly one proof, or at least $k$; \S\ref{sec:construction}).

\paragraph{Shape of the training distribution.}
\citet{zhang2022paradox} show that BERT fits one sampler of propositional Horn problems and fails on
another sampler of the same logic: a generator's statistical shape is learned.

\paragraph{Provenance, landmarks and puzzle generation.}
Mandatory steps are why-provenance \citep{green2007provenance,elhalawati2022provenance} read at the
level of rule applications. In planning they are action landmarks \citep{karpas2009landmarks}, the counterpart for actions of the fact
landmarks of \citet{hoffmann2004landmarks}. Minimal
premise sets are minimal unsatisfiable subsets, enumerated with MARCO \citep{liffiton2016marco}, and
justifications in description logics, found by axiom pinpointing
\citep{baader2010pinpointing,kalyanpur2007justifications,horridge2011justification}. A premise in every
justification is a counterfactual cause of the answer in the sense of database causality
\citep{meliou2010causality}. We carry these notions from premises to rule applications. Puzzle
generators already constrain undesired solutions inside a solver \citep{smith2013quantifying} and generate
natural-deduction exercises \citep{ahmed2013natural}. Our atlas instead lifts such conflicts to schematic
nogoods, proved once per calculus, that prune every later construction.

\section{Realized Structure}
\label{sec:measures}

Consider the problem of Figure~\ref{fig:example}a: premises $a$, $a\to b$, \dots, $e\to f$ and goal $f$,
planted as five applications of modus ponens. If the reasoner may also chain implications (hypothetical
syllogism), adjacent implications can be merged before they are applied, and the goal acquires 42 minimal
proofs. Each still has five applications and uses every premise, but no step belongs to all of them. A
depth label of five merges three quantities that have come apart here: how long the shortest proof is,
which steps all proofs share, and how many proofs there are. This section defines them.

\paragraph{Setting.} Each \emph{logic} in our framework fixes a signature, a calculus of sound rule
schemas, a rendering into English, and a semantics decided by a complete solver (z3 \citep{demoura2008z3},
a CSP solver, or a Datalog chase). Semantics are first-order: quantified and modal statements compile into
decidable fragments (monadic first-order logic, modal logics through the standard translation, finite
domains for puzzles and plans), and a query the solver cannot settle is dropped. Calculi are ground and
forward (no hypothetical steps). A problem is a set of premises $\Gamma$, a hypothesis $h$ and a label:
entailment, contradiction (together E/C) or neutral. Our claims come at three levels, which we keep apart
throughout. Labels are exact relative to the solver. Structure is exact relative to the logic's calculus
over a bounded universe of formulas. Difficulty for models is measured, never certified.

\begin{definition}[Proofs and mandatory steps]
A rule \emph{application} is a rule, a conclusion and premises. A proof of goal $g$ from $\Gamma$ is a
set of applications that derives $g$ from $\Gamma$. An application $s$ is \emph{mandatory} if it occurs in
every proof, i.e.\ $g$ is not derivable once $s$ is removed. Mandatory premises are defined the same way.
\end{definition}

\paragraph{Depth is relative to a calculus.} Every count of proof steps presupposes a set of rules, which
fixes the \emph{granularity} of a step. The planted depth is a count in the generator's own rules. A measure of required structure cannot
avoid this dependence, but it can state its calculus and say how its verdicts move when the calculus
changes. They move in one direction.

\begin{proposition}[Monotonicity]\label{prop:mono}
Let $C\subseteq C'$ be sound calculi over one set of formulas, large enough for both (e.g.\ containing the assumptions of hypothetical steps). Every step mandatory for $g$ under $C'$ is
mandatory under $C$; min depth and min size under $C'$ are at most those under $C$; and every
inclusion-minimal $C$-proof stays inclusion-minimal under $C'$.
\end{proposition}
\noindent Adding rules only adds proofs, so a step some proof avoids stays avoidable (proof in
Appendix~\ref{app:proof}). A finding of \emph{unforced} structure therefore holds for every richer
calculus, with hypothetical reasoning or fused macro-rules: adding rules cannot reverse it. A finding of
\emph{forced} structure is exact for the stated calculus and robust in rank: fused two-step rules lower
counts by a fifth but keep the ranking of problems (Spearman .83; .98 for min depth).

\paragraph{Profile.} Deciding which steps are mandatory takes one closure and one removal test per step
of a single proof. The topology of the mandatory steps is read off the sub-derivation they form: its
\emph{spine} is the longest chain, and its \emph{joins} are the mandatory steps that combine two or more
mandatory sub-derivations. We call the number of mandatory steps the \emph{certified depth} of a problem,
as opposed to its \emph{planted depth}, the length of the proof its generator sampled. It counts steps
that every proof makes, and is neither the height of a shallowest derivation nor the size of a smallest
proof, which the profile reports separately. The \emph{profile} of a goal adds the height of the shallowest derivation (min
depth), the fewest applications in any proof (min size) and the number of inclusion-minimal proofs. When
proof enumeration reaches its cap, the profile is flagged inexact. Min depth and mandatory steps stay
exact (Appendix~\ref{app:details}).

\paragraph{When no step is mandatory.} Two disjoint 10-step proofs give no mandatory step but a min size
of 10, so the absence of mandatory steps does not by itself mean a problem is easy. We split such problems
into four groups: \emph{given} (min size 0), \emph{one step} (min size 1, several rules),
\emph{re-routed} (several proofs of two or more steps that share no step but use the same premises), and
\emph{over-determined} (several minimal premise sets). The last two differ in kind. Re-routing depends on
the granularity of the calculus: the chain of Figure~\ref{fig:example}a has many bracketings and no
mandatory step, though every premise is needed, and without hypothetical syllogism every step of the chain
is mandatory. Over-determination is a property of the premises, the same in any complete calculus. The
four groups partition the problems without a mandatory step: first by min size, then, from two steps on, by
the number of minimal premise sets. We also report the \emph{forced share}: mandatory steps over min size,
pooled over problems with min size at least one.

\paragraph{Premise level.} Without any calculus, the solver alone gives the inclusion-minimal premise sets
that still yield the label (enumerated completely with MARCO): premises needed, alternative
justifications, distractor share, and rule kinds common to every justification. This level applies to any
dataset with logical forms, including those whose rules we do not model.

\section{Auditing Reasoning Datasets}
\label{sec:audit}

We audit each dataset from its own logical forms, translated into one of our logics
(Appendix~\ref{app:details}), and first check the translation: all 11,376 ProofWriter True/False questions
re-derive, and \fld{} labels agree with z3 on all 455 problems drawn. The audit has two parts. The
formal audit (\S\ref{sec:formal}) measures what each problem requires, without any model: the datasets
fall into three cases, claimed depth that is forced, long proofs with mostly optional steps, and answers that need
no particular step (Table~\ref{tab:proof-audit}). The empirical audit (\S\ref{sec:empirical}) asks whether
these measures, or the claimed depth, predict where language models fail.

\begin{table*}[t]
\centering\small
\resizebox{\textwidth}{!}{%
\begin{tabular}{ll rrr rr rrr}
\toprule
& & \multicolumn{3}{c}{depth (steps)} & \multicolumn{2}{c}{necessity} & \multicolumn{3}{c}{problems with no mandatory step} \\
\cmidrule(lr){3-5}\cmidrule(lr){6-7}\cmidrule(lr){8-10}
dataset (sample size) & diagnosis & claimed & shallowest & smallest & mandatory & forced & all & re-routed & over-determined \\
& & & proof & proof & steps & share & & & \\
\midrule
ProofWriter, depth 5 (300) & forced & 5 & 5.00 & 5.05 & 4.74 & 94\% & 0.7\% & 0.3\% & 0.3\% \\
ProofWriter, all depths (1,800) & forced & 2.5 & 2.50 & 2.52 & 2.35 & 93\% & 19\% & 0.1\% & 0.6\% \\
\fld{}, depth 5 (400) & optional & 5 (4.35) & 2.38 & 3.08 & 2.08 & 68\% & 18\% & 17\% & 0\% \\
\fld{}, all depths (400) & optional & 2.0 (1.73) & 1.43 & 1.65 & 0.98 & 60\% & 40\% & 21\% & 0\% \\
ZebraLogic, propagated cells (400) & re-routed & -- & 2.79 & $\le$3.48 & 0.84 & $\ge$22\% & 71\% & 33\% & 6\% \\
PlanBench Blocksworld (501) & forced & -- & 7.6 & 7.6 & 6.3 & 84\% & 0\% & 0\% & 0\% \\
\midrule
Forward generation, propositional (400) & mostly forced & -- & 5.31 & 8.98 & 7.86 & 87\% & 1.5\% & 1.0\% & 0\% \\
Our construction, propositional (400) & forced by design & -- & 7.44 & 13.31 & 13.31 & 100\% & 0\% & 0\% & 0\% \\
\bottomrule
\end{tabular}}
\caption{What problems require, by dataset (means over problems). \emph{Diagnosis}: whether the claimed
depth is forced, mostly optional, or met but by many interchangeable routes. \emph{Shallowest} and \emph{smallest
proof}: the fewest levels and the fewest rule applications of any proof. \emph{Mandatory steps}: rule
applications present in every proof; \emph{forced share}: mandatory steps over smallest proof size. The
last three columns are shares of all problems: \emph{re-routed}, several proofs over one premise set;
\emph{over-determined}, several minimal premise sets; the rest of ``all'' is answers that are given or
take one step. For PlanBench, mandatory steps are action landmarks and the smallest proof is the optimal
plan. In parentheses, \fld{}'s claim without the instantiation steps of its gold proof, which our ground
calculus does not count. \fld{} depth~5 is a separate sample of derivable depth-5 problems (95\% bootstrap
interval for mandatory steps 1.94--2.21). ZebraLogic proof enumeration hits its cap on 46\% of cells, so
its smallest proof is an upper bound and its forced share a lower bound. The last two rows are ours: problems in our propositional logic
under the \fld{} calculus, produced by forward generation (sampling a proof and verbalizing it, as \fld{}
does) and by the construction of \S\ref{sec:construction}.}
\label{tab:proof-audit}
\end{table*}

\begin{figure}[t]
\centering
\includegraphics[width=\columnwidth]{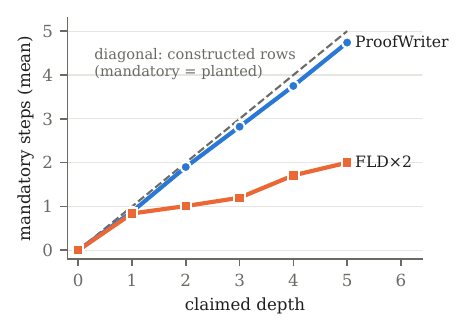}
\caption{Mandatory steps by claimed depth. ProofWriter's claim is nearly forced; \fld{}'s depth-5 problems
force two steps. Constructed problems lie on the diagonal by construction (Table~\ref{tab:proof-audit}).}
\label{fig:audit}
\end{figure}

\subsection{Formal audit: what problems require}
\label{sec:formal}

\paragraph{Depth that is forced: ProofWriter and PlanBench.}
ProofWriter defines depth as the depth of the shallowest proof, so our min depth matching it at every
level (Figure~\ref{fig:audit}) checks the certifier rather than the dataset. The finding is that this depth
is also forced: 93\% of the smallest proof is mandatory. ProofWriter's weaknesses lie elsewhere. 85--90\%
of theory facts are never used, and depth-5 questions draw on only 42 distinct multisets of mandatory rule
kinds. Blocksworld is similar: 84\% of the actions of an optimal plan are landmarks, and 53\% of instances
have a unique optimal plan. PlanBench also offers a natural experiment on what a structural label leaves
out. Mystery Blocksworld renames Blocksworld, and certification confirms that the two are structurally
identical (same lengths, landmarks and plan counts). Reported accuracy of pre-o1 models nevertheless falls
from up to 62.6\% to under 5\% \citep{valmeekam2024strawberry}: with structure held fixed, that drop
measures the wording alone.

\paragraph{Long proofs, mostly optional steps: \fld{}.}
\fld{}'s claimed lengths hold up. Once the instantiation of universal facts is set aside (\fld{} counts
it as a step; our ground calculus does not), the smallest proof matches the claim up to depth~4 and falls
short only at depth~5, where about half the problems admit a shortcut (Table~\ref{tab:fld-depth} in Appendix~\ref{app:details}). Mandatory-step
counts grow much more slowly. A depth-5 problem has about two steps that every proof must take. The
rest can be routed around, and alternative proofs multiply with depth. Two problems in five have no
mandatory step at all, because the hypothesis is already a premise or because the same premises combine in
several orders.

The picture does not depend on our calculus: under \fld{}'s own rules it is the same, and adding
contraposition forces even fewer steps, as Proposition~\ref{prop:mono} predicts
(Appendix~\ref{app:details}). Given hypotheses and the absence of over-determination hold in any
calculus. At the premise level \fld{} is clean but thin: the gold premises are always the unique minimal
support, but a problem needs only two or three of them and three quarters of the context is distractor.
The 16\% of problems our calculus cannot derive (hypothetical proofs, existential statements), 35\% at depth~5, are not counted.

\paragraph{Answers that need no particular step: ZebraLogic.}
ZebraLogic is the opposite case: well-posed puzzles whose individual answers need no particular step. In
an independent check every clue of each parsed grid is needed for the full solution, yet most cells that
propagation reaches have no mandatory step. They are reached by several routes, almost always over the same
clues: only 6\% of cells (3\% of the positive cells ZebraLogic asks for) have a second minimal set of
clues (Table~\ref{tab:proof-audit}). Re-routing depends on the granularity of the calculus
(\S\ref{sec:measures}), so this finding carries the same caveat as \fld{}'s. A model can thus be right about a cell while ignoring most of the
interactions the puzzle was built around. Accuracy on cells says little about whether they were used.

\paragraph{Premise-level audits: FOLIO, ProverQA and LogicGraph.}
Where we do not model a dataset's rules, the premise level still applies. Corrected FOLIO
\citep{han2024folio,fixingfolio2026}, whose quantifiers, exclusive-or and biconditionals fall outside our
calculi, is lean: a label needs 4.1 of 6.6 premises on average, and 93\% of problems have a unique minimal
justification. ProverQA \citep{qi2025provergen} and LogicGraph \citep{wu2026logicgraph} verify their data
with a prover, which guarantees labels and valid proofs. The audit asks whether those proofs are needed.
In ProverQA, often they are not. The premises cited by a problem's reasoning chain always suffice, but a
smaller set already gives the conclusion in 8\% of easy, 30\% of medium and 54\% of hard problems, so the
chain overstates the problem most where it is meant to be hardest. LogicGraph claims an exhaustive set of
minimal proofs per query, and the claim holds where our enumeration is complete (all Small and Medium
instances, a quarter of Large ones): every listed proof's premise set is minimal and no support is missed. Its structure is as claimed. It is also fully redundant at scale, since no Large instance has a
premise that every proof needs. Label agreement and enumeration details are in
Appendix~\ref{app:details}.

\paragraph{State tracking: Boxes.}
The same analysis reaches beyond logic puzzles. The boxes data of \citet{kim2023entity} describe objects
in seven boxes, then put, remove and move operations, and ask what a box contains. Declared as a logic with no change to our machinery (Appendix~\ref{app:details}), it yields two findings
that sharpen the authors' own analysis, which scores accuracy by the number of operations that changed the
asked box. On the base
split, the last sentence naming each object gives the answer exactly, at every length: composition never
matters. And the operations that touch a box overstate those it depends on, since later operations
overwrite earlier ones: with five or more, the answer depends on fewer in 94\% of cases. Moving contents
without naming them, as the authors' harder split does, is what makes tracking necessary, and there the last-mention rule falls from 82\% accuracy with one operation to 54\% with five or more (an
independent count and our logic agree on every sampled story).

\subsection{Empirical audit: what predicts model errors}
\label{sec:empirical}

The formal audit says what a problem requires of an ideal solver. Whether that is what makes problems
hard for language models is an empirical question, which we test on \fld{}, whose claimed and required
depth differ most, and on constructed problems, where they coincide.

\paragraph{What makes \fld{} hard.}
Two LLMs, DeepSeek-V4-Flash \citep{deepseek2026v4flash} and gpt-oss-120b \citep{openai2025gptoss} at low
reasoning effort, label 763 derivable E/C problems of
the \fld{} test set from the dataset's own English, balanced over claimed lengths. Almost every error is
``neutral'': the model fails to find a derivation, rather than finding a wrong one. Errors grow with
claimed length, as a depth label promises, but length is not the only thing that grows. Longer \fld{}
proofs more often use a universal fact, and at a fixed claimed length such a fact costs gpt-oss 20 to 40
points of accuracy. In joint logistic fits, universal facts predict errors beyond every depth measure for
both models, and the depth measures add information beyond universal facts as well
(Table~\ref{tab:llm-depth} in Appendix~\ref{app:details}). The depth label thus leaves out a large factor that it does not measure.

The models' replies suggest why. \fld{} words most universals with ``some'', ``somebody'' or
``something'' (``Some man is unplanned if the one is a kind of a shadowed''), and the replies read them as
particular (``It does not state that every shadowed thing is unplanned''). Rewording tests whether this is
the cause. On the 161 problems whose gold proof uses such a fact, we change only the quantifier word
(``Something that\dots'' becomes ``Everything that\dots''; the logic is untouched) and ask each model
three times per version. Comparing replies obtained on the same day, rewording raises gpt-oss's accuracy by
7 points (.59 to .66, interval over problems .00--.13) and DeepSeek's by 5 (.58 to .63, $-$.01 to .11). The
wording is thus one cause of errors, borderline for each model. (Replies to the unchanged prompts from an
earlier run differ from same-day replies by as much as this effect, so only same-day replies are compared.)
Necessity, unlike length, does carry signal: for
gpt-oss, problems whose smallest proof is fully forced are easier and problems whose premises can be
re-routed harder (forced share +0.47 [0.28, 0.68] in a fit with claimed length, min size and universal
facts; no effect detected for DeepSeek). As with Mystery Blocksworld, part of what predicts errors is
something the depth label does not measure.

\paragraph{Mandatory steps predict errors.}
Constructed problems (\S\ref{sec:construction}) remove both confounds: every planted step is mandatory,
the planted proof is the shortest, and universals are always worded ``every'', ``all'' or ``no''. On such
problems depth does what \fld{}'s label could not (Table~\ref{tab:llm-constructed}, 600 monadic problems
with 1--20 mandatory steps). gpt-oss-120b goes from near-perfect on short proofs to near-floor on long
ones, and more reasoning effort barely moves it. Gemma-4-31B \citep{gemma2026gemma4} is near ceiling, but its few errors sit at
the deepest problems. DeepSeek-V4-Flash is perfect throughout, so the problems are solvable as written.
The failure is again ``neutral'': the model reports a missing premise that, in every error we traced by
hand, is in the text. Depth is not standing in for length here: at a fixed number of premises, accuracy
still falls with depth (below 15 premises, from .58 up to 7 steps to .17 from 13).

\begin{table}[t]
\centering\small
\resizebox{\columnwidth}{!}{%
\begin{tabular}{lcccc}
\toprule
mandatory steps & 1--3 & 4--7 & 8--12 & 13+ \\
\midrule
DeepSeek-V4-Flash (low) & 1.00 & 1.00 & 1.00 & 1.00 \\
gpt-oss-120b (low) & 0.89 & 0.40 & 0.13 & 0.10 \\
gpt-oss-120b (high) & 0.90 & 0.45 & 0.22 & 0.13 \\
Gemma-4-31B & 1.00 & 1.00 & 0.99 & 0.95 \\
\midrule
problems (E/C) & 29 & 73 & 139 & 166 \\
\bottomrule
\end{tabular}}
\caption{Accuracy on constructed monadic E/C problems by certified mandatory depth. Neutral problems: 0.97--0.99.
gpt-oss (high), standardized logistic coefficients: mandatory steps $-0.63$ [$-1.00$, $-0.33$], premises
$-0.66$ [$-1.07$, $-0.25$]; Gemma: mandatory $-2.29$ [$-5.67$, $-1.32$], premises $+0.48$ [$-0.29$, $1.34$]. Gemma makes about ten errors in all, so its coefficients are imprecise.}
\label{tab:llm-constructed}
\end{table}

\section{Constructing Certified Structure}
\label{sec:construction}

The audit separates planted from realized structure after the fact. Construction closes the gap
beforehand: a request states the structure every proof should have, and the output is a problem certified
to have it. A \emph{target} requests a label, a multiset of rule applications or the shape of the mandatory steps (spine and joins), a
number of minimal proofs, and distractors (irrelevant derivations and near misses that miss the goal by
one premise). Our construction algorithm, \proofplant{} (Algorithm~\ref{alg:proofplant}), builds an
intended derivation backward and binds it to fresh constants. It then certifies the problem: every
planned application must be mandatory, and the topology of the mandatory steps must equal the requested
shape; the rest of a proof may vary.

\begin{figure*}[t]
\centering\small
\renewcommand{\arraystretch}{1.15}
\begin{tabular}{@{}p{0.15\textwidth}p{0.38\textwidth}p{0.41\textwidth}@{}}
\toprule
& \textbf{(a) Planted by forward generation} & \textbf{(b) Constructed by \proofplant{}} \\
\midrule
premises & $a,\ a{\to}b,\ b{\to}c,\ c{\to}d,\ d{\to}e,\ e{\to}f$
  & $n\vee k,\ j,\ t{\to}g,\ \neg g,\ j{\to}(\neg t{\to}(\neg n\wedge b))$ \\
& & \emph{Either Nora is happy or Kenji is calm. Jonas studies law. If Tara drinks tea, then Grace is
  busy. Grace is not busy. If Jonas studies law, then (if Tara does not drink tea, then both Nora is not
  happy and Bob is tired).} \\
goal & $f$ & $k$ (\emph{Kenji is calm}) \\
planned proof & MP, MP, MP, MP, MP: $a \Rightarrow b \Rightarrow \dots \Rightarrow f$
  & MT: $\neg t$; MP: $\neg t{\to}(\neg n\wedge b)$; MP: $\neg n\wedge b$; $\wedge$E: $\neg n$; DS: $k$ \\
\midrule
minimal proofs & \textbf{42} (with hypothetical syllogism, HS) & exactly \textbf{one} \\
mandatory steps & \textbf{0 of 5}: every MP can be routed around (re-routed, not over-determined)
  & \textbf{5 of 5}; spine 4, one join \\
why & nogood $\textit{mp}(\textit{mp},\_)$: HS bypasses an MP fed by another MP
  & no placement completes a nogood \\
\bottomrule
\end{tabular}
\caption{Planted vs.\ certified structure in propositional logic, row by row. (a) A planted five-step
chain in which no step is mandatory once hypothetical syllogism is available. (b) A problem built by
\proofplant{} for the plan \{MP, MP, MT, $\wedge$E, DS\}, whose five planned steps are all mandatory;
its certificate lists the proof and its mandatory steps, and an independent checker re-derives both.}
\label{fig:example}
\end{figure*}

\begin{algorithm}[t]
\caption{\proofplant{}: atlas-guided construction of a certified problem.}
\label{alg:proofplant}
\small
\begin{algorithmic}[1]
\Require target $T$ (label; plan: rule multiset or mandatory-step shape; $k$ minimal proofs;
  distractors), logic $P$, atlas $A$ of nogoods
\Ensure a problem $(\Gamma, h)$, its label and its certificate, or failure
\For{attempt $= 1, \dots, N$}
  \State $D \gets$ derivation of the goal built backward from $T$'s plan: place a rule schema at an
    open node by unification, \textbf{unless} the placement completes a nogood of $A$ (constants kept
    distinct)
  \State bind the metavariables of $D$ to fresh constants; add $T$'s distractors
    \Comment{gives $(\Gamma, h)$}
  \State $\ell \gets$ label of $(\Gamma, h)$ from $P$'s solver
  \State $\pi \gets$ profile of $(\Gamma, h)$ \Comment{mandatory steps, topology, minimal proofs}
  \If{the requested steps are mandatory in $\pi$ (every planned step; for $k \ge 2$, the requested
    bottlenecks and at least $k$ minimal proofs) \textbf{and} $\pi$'s topology is $T$'s}\label{alg:accept}
    \State \Return $(\Gamma, h)$, $\ell$ and the certificate (proof, mandatory steps; countermodels if
      neutral)
  \EndIf
  \State lift the fragment of $D$ that a proof bypasses to a schematic nogood $n$; verify $n$ over
    $U^*(n)$ and add it to $A$ \label{alg:learn}
\EndFor
\State \Return failure
\end{algorithmic}
\end{algorithm}

\paragraph{Why filtering is not enough.}
The obvious alternative is to generate forward and keep the problems that certify. In a redundant calculus
this fails more often as size grows, because deleting distractors cannot remove a shortcut that lies
\emph{inside} the intended proof. At 30 rule applications, only 5\% (propositional) and 15\% (monadic) of
forward-generated problems need all their planned steps. Backward construction with per-instance
rejection fails 33--44\% of requests when it may redraw each one eight times, as the atlas may; given 64
redraws it succeeds on 98--99\%, and the atlas still makes a certified instance two to three and a half
times cheaper (Table~\ref{tab:baselines} in Appendix~\ref{app:details}).

\paragraph{Shortcut atlas.}
The shortcuts that defeat planting are few and recurrent. In Figure~\ref{fig:example}a the cause is the
same at every step: a modus ponens whose minor premise comes from another modus ponens is bypassed by
hypothetical syllogism, whatever the formulas involved. The atlas records such patterns once per calculus,
and construction never builds them.

Formally, a \emph{fragment} is the top of a derivation tree: a rule, and for each premise either a hole or
a sub-fragment. Its most general instances (MGIs) apply the schemas with most general unifiers and replace
the remaining metavariables by fresh, distinct constants. A fragment $F$ is a \emph{nogood} if every MGI
has a proof of its root that avoids one of $F$'s applications, checked over a stable universe $U^*$ of
subformulas, their non-cancelling negations and closed constructors.

\begin{theorem}[Soundness]
Let $F$ be a nogood and $D$ a tree derivation in which no application occurs twice, in an instance that
may contain any further premises and distractors.
If $D$ contains $\sigma(M)$ for an MGI $M$ of $F$ and a sort-preserving substitution $\sigma$ that is
injective on $U^*(M)$ and maps it into the instance's universe, then some proof of $D$'s conclusion avoids
one of $D$'s planned applications.
\end{theorem}
\noindent The proof (Appendix~\ref{app:proof}, which also shows both hypotheses necessary) maps the
shortcut of $M$ into the instance and splices it into $D$. Atlases enumerate fragments up to three
applications (about a second per logic) and are small: the propositional one has six nogoods, e.g.\
$\textit{mp}(\textit{mp},\_)$, shortcut by hypothetical syllogism. Construction rejects a placement as
soon as it completes a nogood. Every output is still certified in full, so pruning can only lose valid
problems (Appendix~\ref{app:proof}). When certification still finds a shortcut, the responsible fragment
is lifted to a new schematic nogood, which prunes all later constructions of any size.

The atlas also makes explicit what a calculus can teach. With hypothetical syllogism among the rules, no
constructed problem contains a chain of modus ponens, because no such chain can be forced there. Which
chains appear follows from the calculus a logic declares. A logic that should teach such chains omits
hypothetical syllogism.

\paragraph{Scale and coverage.}
With the atlas, failure by shortcut disappears. Certified success is 99--100\% for propositional and
monadic requests of 20, 30, 40 and 60 steps (200 requests each; 0.03--0.34\,s per problem). Construction
also steers which proof shapes appear. It keeps an archive of shape classes (label $\times$ spine length
$\times$ number of joins, spine up to 12, joins up to 6), picks the least-covered class, draws a skeleton
uniformly among the trees of that class, and fills it. With 380 E/C problems per strategy, this reaches
all 140 E/C shape classes in each of five logics (propositional, monadic, CSP, modal K, order), against
78--88 for forward generation, at 0.75--1.5$\times$ its throughput. The lead survives size matching, and
the same archive can follow a requested distribution over classes instead of uniform coverage
(Appendix~\ref{app:details}).

\section{Certificates and Logics}
\label{sec:certificates}

\paragraph{Independent checking.}
A structural claim is only as good as the code that computes it, so each problem exports a certificate
that can be checked without that code: the proof, the claimed mandatory steps and premises, the background
facts, and the label query in SMT-LIB. The checker trusts what the certificate refers to: the rules of the
calculus and the logic's encoding into SMT, whose soundness is checked separately (\S\ref{sec:certificates},
Logics). A checker that shares no code with the generator re-verifies each
proof step and re-saturates without each one: the claimed mandatory steps must be exactly those whose
removal makes the goal unreachable. A neutral label carries finite \emph{countermodels}, which the checker
evaluates directly, without a solver. Over 540 certificates from 9 logics, all pass, and every injected
corruption that invalidates a certificate is caught (Appendix~\ref{app:details}). Certificates depend only
on the problem as shown: analysed from its premises and hypothesis alone, without the formulas the generator
passed through, each of 800 generated problems (four logics, both generators) has the same depth,
mandatory steps and smallest proof.

\paragraph{Logics.}
Fifteen logics are instantiated: propositional logic, monadic first-order logic (with and without
syllogisms), a CSP propagation calculus, modal K and S4 (flat and nested), hybrid modal logic with named worlds, strict orders, Zebra-style grids, a kinship
program imported from Datalog, bounded planning, objects moved between boxes, and arithmetic over scores. Each lists its signature,
rule schemas, semantics and rendering, and its rules are proved sound for all instances when it is loaded
(137 rules in all, each proved for every instance by z3). Logics are also checked for
\emph{requiredness}: whether each rule is mandatory in some instance, since structural claims are only
informative for calculi in which each rule can be required. Every rule passes except two, kept on purpose
as one-rule shortcuts: in flat modal S4, ``necessarily $A$ at $w$, and $w$ sees $v$, so $A$ at $v$''
chains two other rules, and in arithmetic, ``$P$ scores below $Q$, so $P$ is not higher than $Q$'' does
too. They are never mandatory, so construction never plans them as steps that must be made. What a logic can express, and where
construction stops, is detailed in Appendix~\ref{app:details}.

\section{Training with Certified Structure}
\label{sec:transfer}

\begin{figure*}[t]
\centering
\includegraphics[width=\textwidth]{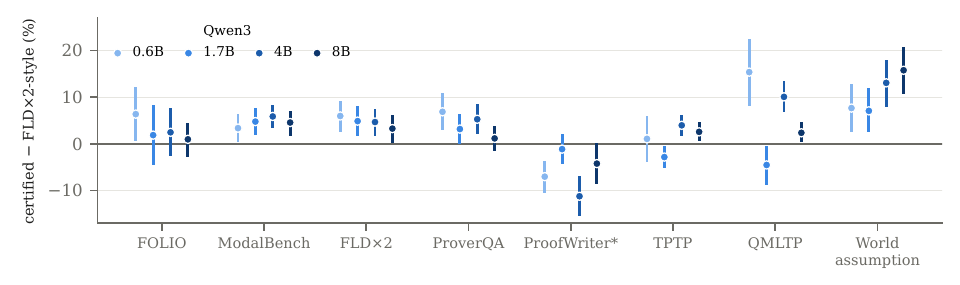}
\caption{Gain from training on certified data instead of \fld{}-style data. Each point is the two-way
accuracy of a model trained on the certified mixture minus that of the same model trained on \fld{}-style
examples from the same logics, in percentage points; above zero, certified data is better. Both training
sets have 3{,}000 examples; points are means over seeds 10--16 with 95\% intervals. $^\ast$ProofWriter's
labels follow a surface cue (\S\ref{sec:transfer}). All scores: Table~\ref{tab:search}.}
\label{fig:forest}
\end{figure*}

Construction makes depth something we can guarantee. This section asks how training data should use it:
whether certified depth changes what a model learns, whether more of it is better, and how depth should be
combined with other properties of the data. Certification is what makes these questions testable, because
training sets can be made to differ in one structural variable while labels, premise counts and neutrals
are matched.

\paragraph{Setup.} We fine-tune small LLMs, Qwen3-0.6B-Base \citep{yang2025qwen3} unless stated, to predict the label (full
fine-tuning, 2 epochs, 3{,}000 examples) and score labels by log-probability. Label prediction is the
simplest training signal, and it keeps differences between training sets attributable to the data rather
than to a decoding or reward scheme. The experiments therefore show how certified structure shapes label
prediction and transfer. Models can also be trained to write proofs, which our checkers grade: trained on
certified problems, they answer deep problems while citing premises that suffice for the answer several
times more often than when trained on forward problems in the same format, and the gap opens with depth (Figure~\ref{fig:proofs} in
Appendix~\ref{app:llm}). Encoders point the same way and are left to
Appendix~\ref{app:encoders}: constructed data leads \fld{}'s own training data on its test by 1--2 points,
and rule kinds and shapes matter more than depth.

\paragraph{Controls and metric.} Comparisons between reasoning datasets are easily decided by something
other than reasoning. In our first comparisons, shared neutral examples, generator style, a
hypothesis-only cue and three-way accuracy on test sets without neutral problems each decided the outcome
until a matched control removed it (Appendix~\ref{app:shortcuts}). Training sets are therefore matched on
labels, premise counts and how neutrals are built, and we report \emph{two-way accuracy} (on problems labelled E or C, whether the model prefers the
right one of these two labels, balanced over both) and the neutral rate separately. Two-way accuracy
cannot be raised by answering neutral more often, since neutral is not an option.

\paragraph{Certified depth alone changes little.} The audit found that \fld{}'s depth-5 problems force two
steps under standard derived rules. Whether such a gap matters for training can be tested directly. We take
forward-generated monadic examples planted at depth 5--9 and split them by their number of mandatory
steps: at most three (the \fld{} pattern), or at least the planted depth minus one. The two training sets are
matched per label and planted depth (mean 5.8; certified 2.5 vs.\ 5.8) and in length (472 vs.\ 490
characters), and their neutrals are split the same way by the depth of the proof their deleted premise
belonged to. A first version of this contrast seemed to show that certified-deep examples train better
(two-way accuracy +.11 on QMLTP, +.03 on \fld{} and ModalBench) and make the model abstain far more
often. Most of both effects came from a label shortcut that selecting by certified depth brings along:
deep proofs more often end in a negated contradiction, so negation separated contradictions from
entailments by .34 in the deep set against .13 in the shallow one, and the deep set gained where a
negated hypothesis signals contradiction and lost where it misleads (Appendix~\ref{app:polarity}). With
negation balanced in both sets (half the examples of each label and planted depth negated), fourteen seeds
of Qwen3-0.6B show no two-way difference on any test (largest: QMLTP +.05, interval $-$.00 to .10), and
more neutral answers only on FOLIO, ProverQA and ProofWriter. At a fixed planted depth, then, what
certification changes in this data is mostly which conclusions proofs reach, not how well a model trained
on it reasons. Full results for this and the other
controlled contrasts are in Table~\ref{tab:ablations} (Appendix~\ref{app:llm}).

\paragraph{Deep examples need short ones beside them.} We split examples into three ranges of mandatory
steps (1--3, 4--7, 8+), matched on label and premise count; a neutral example deletes one
premise of a proof and is placed by the depth of that proof. Deep-only training buys little on deep test
examples and costs the short ones, and on \fld{} it fails in a telling way (Table~\ref{tab:llm-mix},
Appendix~\ref{app:llm}). Qwen trained on deep examples alone answers neutral on half of \fld{}'s derivable
problems, yet the answers it does commit to are as accurate as those of any other mixture. Deep-only
training makes models ready to declare a label underivable, and short proofs hold that back. Every mixture with shallow examples avoids this collapse, in three model
families, while finer rankings between such mixtures do not carry over between models
(Figure~\ref{fig:mix}). Distractor load contributes too: depth ranges are matched on premise count, so
shallow examples carry more irrelevant premises, and without any distractors shallow-only training reaches
.54 on \fld{} instead of .62, still well above deep-only data. Training data should cover a range of depths
that includes short proofs, which is also what the benchmarks we transfer to look like.

\paragraph{Redundant proofs are not a defect.} The audit also finds problems whose goal can be reached in
many ways from the same premises. Training on such problems (several minimal proofs, three or more for 71\% of them, re-routings
of one premise set) instead of single-proof problems with the same shortest proof changes two-way accuracy
by at most three points, on two of eight tests and in opposite directions (fourteen seeds,
Table~\ref{tab:ablations}). How deep a problem's proofs are matters for training, but how many routes it
offers matters little.

\paragraph{Sources.} Depth is one axis of the distribution; the kind of problem is another, and certified
sources can be mixed. We add two to the monadic examples: labelled modal logic problems, and closed Kripke
models where the question is whether a formula holds at a world (z3, checked by an independent evaluator;
E/C only). Mixtures selected on development sets (\fld{} validation, 400 FOLIO training examples, a third
of ModalBench's English track~\citep{hasan2026modalbench}) and confirmed on fresh seeds improve FOLIO and
ModalBench and leave \fld{} unchanged (Table~\ref{tab:sources} in Appendix~\ref{app:llm}). The gains survive when every training set
is restricted to E/C examples, so they come from what the sources teach and not from mixing as such.

One thing they teach is wrong, and the way it is repaired shows what control over the distribution is for.
A Kripke example lists every fact of its world, and the model carries this closed-world reading to text
that does not license it: FOLIO neutrals start being answered as contradictions. Hiding a share of the
Kripke facts, so that some formulas are undecided and labelled neutral, undoes this while keeping the
other gains (designed after the tests exposed the problem, then confirmed on fresh seeds). It also teaches
the model to read the stated assumption, and one failure remains in every mixture: models keep answering
entailment when a mandatory premise is removed (Appendix~\ref{app:llm}).

\paragraph{Mixtures of certified problem types.} No single depth range or source is best, so we describe
the training distribution by nine certified problem types (monadic and modal examples in three depth
ranges, closed Kripke examples by modal depth, partially observed Kripke examples) and mix them. At equal
size we compare mixtures over these types (uniform, or searched as described below) with an \fld{}-style
baseline (forward-generated monadic and modal examples, half each, labelled by the same solver and
rendered by the same templates, taken as the generator makes them, with no selection on certified
structure; forward generation has no counterpart for Kripke examples) and with 3{,}000 training examples
of \fld{}, ProverQA and ProofWriter. \fld{}-style generation is the main baseline because it is the
established way to build deductive training data: FLD and \fld{} were designed as training
corpora and shown to improve LLM reasoning \citep{morishita2023fld,morishita2024fldx2}, and ProverQA
also builds each problem from a sampled proof. Rebuilding it in our own logics, with our solver and templates,
leaves the two training sets differing in whether their structure is certified and chosen, and through that
choice in their composition of problem types and depths. We separate the two below. Besides the five tests used so far, three were seen by no selection
step: TPTP~\citep{sutcliffe2017tptp} and QMLTP~\citep{raths2012qmltp} problems rendered in English with
prover-checked labels (1{,}412 and 900 examples), and 300 world-assumption pairs, Kripke problems that
differ only in the sentence closing the world (Table~\ref{tab:search}).

Certified problem types beat \fld{}-style examples on most tests, at every model size we tried
(Figure~\ref{fig:forest}). The lead
is systematic on \fld{}, ModalBench and ProverQA, where it holds in all five Qwen settings (ProverQA
except at 8B) with paired bootstrap intervals over test examples that exclude zero, and uniform weights
over the types already obtain it. It also holds on the world-assumption pairs, partly by construction,
since only the certified mixture contains Kripke examples. \fld{}-style examples win on ProofWriter at
every size, mostly through a surface cue: 96\% of its test contradictions and 7\% of entailments have a
negated hypothesis, so answering contradiction exactly on negation scores .945. On the 22 problems where
the cue misleads, the certified mixture leads instead (by .17 at 0.6B and .10 at 8B), as it does on
TPTP's (Appendix~\ref{app:polarity}). On the prover-derived tests the
picture turns with scale, \fld{}-style examples leading at 1.7B and the certified mixture from 4B on,
which suggests that the harder problem types pay off once the model can use them. The gains transfer only
partially to a second model family: SmolLM2-1.7B \citep{allal2025smollm2}, at chance on ModalBench and QMLTP whatever it is trained on, keeps the TPTP lead
and the neutral rate. The neutral rate is the most portable effect: the certified mixture's stays
near the gold share across tests, while existing datasets carry their label prior with them: on
ModalBench, where no example is neutral, they answer neutral on most examples. Against those datasets, the
certified mixture is the best out-of-domain source on FOLIO and ModalBench and close to the best
elsewhere, except on ProofWriter and TPTP.

\paragraph{Where the lead comes from.} The comparison above changes two things at once, the examples and
their proportions, and the two can be separated. Keeping a searched mixture and swapping only its
derivation examples for forward examples changes accuracy significantly on no test over seven seeds
(Appendix~\ref{app:llm}): at the level of a mixture, the gain is composition. The per-example value of a
certificate is the planted-against-certified contrast above, where composition is held fixed. The two
results fit together: the mixture's problem types are defined by certified depth and include model-checked
examples (neutrals with countermodels, Kripke examples) that a forward generator does not produce.

The same holds for the bins themselves. Building the nine types from forward examples binned by their
planted depth, as a generator without certificates would, and reusing the searched weights, loses on
ModalBench in every seed and slightly on \fld{}, and matches the certified mixture elsewhere; with uniform
weights the two binnings cannot be told apart (Table~\ref{tab:search}). Which depth labels the bins carry
matters only once weights are fitted to them: a search learns how much of each depth to include, and
planted depth misreports what each bin holds.

\paragraph{Searching the mixture.} With problem types as coordinates the mixture can be searched. Following
RegMix~\citep{liu2025regmix}, we train 60 one-seed proxies on random mixtures, regress the mean
development score on the weights, and confirm the predicted optimum on fresh seeds
(Appendix~\ref{app:llm}). Composition matters and is predictable, and the fit reads as an analysis: shallow
monadic examples and deep modal examples help every task and deep monadic examples get no weight, so the
useful control is over \emph{which} depth, per source, and not over depth as such. The selected mixture
beats uniform weights on most tests in the setting it was searched for, without shifting the label prior.
Carried to other model sizes and to 9{,}000 examples, it is mostly level with them. Search is thus an
instrument to rerun at the size of interest, and the lead over \fld{}-style examples does not depend on
it.

\section{Conclusion}

Generators plant depth, and what they plant can be checked. Measured by what every proof requires,
ProofWriter's depth is real and forced, \fld{}'s has its length but is rarely forced, most ZebraLogic grid
cells need no particular step, Blocksworld plans are almost fully forced, and ProverQA's chains are often
longer than needed. A shortcut atlas with a soundness theorem makes the converse practical: data whose
every proof makes the requested steps, with checkable certificates. The training study then qualifies the
goal. At a fixed planted depth, certified depth alone changes little once the label shortcut it brings
along is balanced, and the best training data covers short proofs beside deep ones and several kinds of
problems. Its composition accounts for most of its lead over
uncertified \fld{}-style data from the same calculi. The gap between planted and required structure widens
with depth, so we expect certification to matter more as generators are asked for harder problems.

\section*{Limitations}
\paragraph{Structure is relative to a calculus.} Mandatory structure is defined relative to a set of
inference rules. This is inherent to the notion: a proof step can only be counted once the rules that
license it are fixed, and a different reasoner may take shortcuts that ours cannot. We address this with
generality rather than with a single privileged calculus. A logic is declared by its rule schemas, fifteen
are instantiated (propositional, first-order, modal, constraint, planning and state-tracking reasoning),
and a Datalog program can be imported as one, so a reader who prefers other rules can state them and rerun
every audit. The verdicts then move in a known direction. A step found avoidable stays avoidable under any
sound extension of the calculus (Proposition~\ref{prop:mono}). Forced steps, including those of certified
construction, are exact only for the stated calculus: under coarser rules their counts shrink, although
rankings between problems are largely preserved (from 11.1 to 8.7 mandatory steps on average with fused
rules; topology is the least stable measure, Spearman .66).

\paragraph{What the audits cannot see.} Our logics omit hypothetical reasoning and case splits, so 16\%
of \fld{} problems, FOLIO at the proof level and some ZebraLogic cells are out of reach. We report them
as such rather than approximating them. Proof enumeration is capped, so for the few problems whose
profile is inexact (flagged in the data), the smallest proof size and the number of proofs are bounds.
Each audit also depends on a translation of the dataset's logical forms into one of our logics. We check
these translations by re-deriving every label we can (all ProofWriter True/False questions, every \fld{}
problem drawn), but a translation that agrees on labels could still differ in which rules it exposes.

\paragraph{What construction selects.} Requiring every planned step to be mandatory also decides which
problems can exist. Where the rules offer two routes (with a rule that chains ``all A are B'' and ``all B
are C'', the facts $A(a)$, all A are B and all B are C have two two-step proofs and no mandatory step),
redundant and compositional instances are excluded, so our deep problems are also problems with a single
proof. At a fixed proof length this matters little for training (\S\ref{sec:transfer}), but we have not
tested it with deeper or deliberately over-determined problems. A neutral example's counterfactual depth
is the depth of the proof its removed premise would complete, which does not measure how hard it is to
establish that the problem is neutral. Rule-kind vocabularies differ across sources, so diversity counts
across sources are approximate, while counts within one vocabulary are exact. Finally, our texts are
templated: a certificate covers the formal instance, not the faithfulness or naturalness of its English
rendering, and models trained on templates may learn surface regularities that natural text lacks.

\paragraph{How far the training evidence goes.} The training study uses small models (Qwen3-0.6B with
full fine-tuning, 360M and 270M models, and Qwen3 1.7B to 8B with LoRA \citep{hu2022lora} for the mixture comparison), short
training runs and three to fourteen seeds per setting. Most effects are a few points, small relative to
seed variance, so we report intervals over seeds and test problems and say where model sizes disagree; the
encoder experiments are smaller still. Several external tests carry surface cues of their own (ProofWriter
and TPTP labels follow hypothesis negation), so we split results on them by whether the cue gives the
right answer. Training supervises the label only. Whether the same conclusions about data design hold
when models are trained to write out their reasoning, or with verifiable rewards, is open. A first test
with proof supervision is in Appendix~\ref{app:llm}.

\section*{Ethics Statement}
We release synthetic data and audits of public datasets. No personal data is involved. The audits
report properties of datasets, not of their authors' intentions, and several of them (ProofWriter's
depth, ZebraLogic's clue minimality) confirm the datasets' own claims.

\ifcameraready
\section*{Acknowledgments}
This work was (partially) funded by the National Research Agency under the France 2030 program, with the
reference ANR-25-APEV-0001.
\fi

\bibliography{refs}

\appendix
\section{Proofs}
\label{app:proof}

\paragraph{Proposition~\ref{prop:mono} (monotonicity).} Every $C$-proof is a $C'$-proof: the intersection
of the proofs shrinks, minima drop, and a proper subset of a $C$-proof uses only $C$-applications, so it
proves $g$ under $C'$ only if it does under $C$. $\square$

\paragraph{Soundness theorem.}

\paragraph{Setting.} Formulas are terms over the logic's signature; atoms apply predicates to constants
of declared sorts. A rule schema has premise patterns and a conclusion pattern over metavariables; an
\emph{application} $(r, c, \vec p)$ is a rule with a ground conclusion and premises that match its
schema, with unordered argument groups matched modulo permutation. \emph{negate}$(f)$ is $g$ if
$f=\neg g$ and $\neg f$ otherwise. For a set of formulas $X$, let $L(X)$ be the subformulas of $X$
together with their negations and the double negations $\neg\neg g$ of their negated members $\neg g$
(which modus tollens on a negated antecedent concludes), and let $U(X)$ contain $L(X)$, every closed constructor applied to members
of $L(X)$, and every formula whose negation is such a construction. A proof of $g$ from $\Gamma$ in
universe $U$ is a set of applications whose conclusions and premises lie in $U$ and which derives $g$
from $\Gamma$. An instance with premises $\Gamma$ and hypothesis $h$ is analyzed in $U(\Gamma\cup\{h\})$.

An MGI $M$ of a fragment $F$ instantiates $F$'s schemas with most general unifiers; its remaining entity
metavariables become fresh, pairwise distinct constants and its formula metavariables fresh
\emph{schematic atoms} $S$. Its \emph{stable universe} $U^*(M)$ is defined like $U$, except that $L^*$
omits the negations of schematic atoms. $F$ is a nogood if, for every MGI $M$ with root $g_M$, some proof
$P_M$ of $g_M$ from $M$'s leaves in $U^*(M)$ omits an application $s_M$ of $M$. A substitution $\sigma$
maps constants to constants of the same sort and schematic atoms to formulas, and acts homomorphically.

\paragraph{Lemma 1 (substitution).} If $(r,c,\vec p)$ is an application of $r$, so is
$(r,\sigma c,\sigma\vec p)$.
\emph{Proof.} If $\theta$ matches the schema to $(c,\vec p)$, then $\sigma\circ\theta$ matches it to
$(\sigma c,\sigma\vec p)$; a permutation of an unordered group commutes with $\sigma$. \qed

\paragraph{Lemma 2 (stability).} $\sigma(L^*(M)) \subseteq L(\sigma M)$ and $\sigma(U^*(M))\subseteq
U(\sigma M)$.
\emph{Proof.} $\sigma$ maps subformulas of $M$ to subformulas of $\sigma M$. Let $n=\textit{negate}(f)$
with $f$ a subformula of $M$ and $f\notin S$. If $f=\neg g$, then $n=g$ and $\sigma g$ is a subformula
of $\sigma M$. Otherwise $f$'s head is a predicate, constant connective or symbol other than $\neg$;
$\sigma$ preserves heads of non-schematic formulas, so $\sigma n=\neg\sigma f=\textit{negate}(\sigma f)
\in L(\sigma M)$. A closed construction $k(\vec x)$ over $L^*(M)$ maps to $k(\sigma\vec x)$ over
$L(\sigma M)$, and a formula whose negation is a construction maps to one whose negation is the image
construction (same case analysis on its head). A double negation $\neg f$ with $f=\neg g$ a subformula
of $M$ maps to $\neg\sigma f$, the double negation of the negated subformula $\sigma f=\neg\sigma g$ of
$\sigma M$. \qed

Without double negations in $L$, excluding the negations of schematic atoms was necessary: for $b\in S$
with $\sigma b=\neg q$, $\sigma(\neg b)=\neg\neg q$ was outside the instance's universe, and the
propositional atlas built with $U$ instead of $U^*$ contained a nogood, $\textit{mt}(\textit{hs},\_)$,
that $A{\to}\neg b,\ \neg b{\to}C,\ \neg C\vdash\neg A$ refuted. With them, $\neg\neg q\in L(\sigma M)$
since $\sigma b=\neg q$ is a subformula of $\sigma M$, that problem has a second proof (two modus tollens
steps through $\neg\neg b$), and the exclusion is kept only as a precaution: a smaller $U^*$ can remove
nogoods, never add them.

\paragraph{Proof of the theorem.} Let $F' = \sigma(M)$ be the occurrence in $D$, $P=P_M$ and $s=s_M$.
By Lemma~1, $\sigma P$ (applications mapped one by one) is a set of applications deriving
$\sigma g_M$, the conclusion of $F'$, from $\sigma$ of $M$'s leaves, which are the leaves of $F'$. By
hypothesis its formulas lie in the instance's universe (Lemma~2 gives $U(\sigma M)$; the hypothesis
places these formulas in $U(\Gamma\cup\{h\})$, which holds in particular when the formulas of $F'$ are
subformulas of $\Gamma\cup\{h\}$ or their negations). Since $D$ is a tree in which no application occurs
twice, the internal nodes of $F'$ are used only inside $F'$ and no application of $F'$ occurs elsewhere in
$D$: removing $F'$'s applications from $D$ and adding $\sigma P$ yields a set $Q$ of
applications that still derives $D$'s conclusion from $\Gamma$ (proofs are monotone: extra premises and
distractors add applications but remove none). It remains to show that $\sigma s \notin Q$. By the non-repetition hypothesis, applications of $D$
outside $F'$ differ from those of $F'$, and $\sigma s$ is an application of $F'$. If $\sigma t =
\sigma s$ for some $t\in P$, then $t$ and $s$ have the same rule and their conclusions and premises have
equal images; all of them lie in $U^*(M)$, where $\sigma$ is injective, so $t=s$, contradicting $s\notin
P$. Hence $Q$ is a proof of $D$'s conclusion without the planned application $\sigma s$. \qed

\paragraph{Remarks.} (i) Injectivity is needed: a stress test of the monadic atlas finds 6
counterexamples among 5{,}219 substitutions that identify constants (\textit{celarent} chained with two
classes merged) and none among about 13{,}000 injective ones. (ii) Construction checks injectivity on
constants, which implies injectivity on $U^*(M)$ when schematic atoms are bound to distinct atoms; it does
not check either hypothesis for compound bindings. (iii) Non-repetition is needed: if $\sigma s$ also occurred in another branch of $D$, splicing would leave
it in $Q$. Construction applies nogoods without checking it, which can only reject placements that would
have been valid, never admit an invalid instance. (iv) The theorem only justifies \emph{pruning}.
Every constructed instance is certified in full afterwards, so a nogood applied outside the theorem's
hypotheses can only reject a placement that would have been valid; it cannot produce a wrong certificate.
Guarded nogoods (a fragment bypassed only when a formula metavariable is not a negation) split each
formula metavariable into a plain case and a $\neg w$ case, both closed under substitution, and apply
the theorem per case.

\section{Audit and Construction Details}
\label{app:details}

\begin{table}[t]
\centering\small
\resizebox{\columnwidth}{!}{%
\begin{tabular}{lcc}
\toprule
& DeepSeek-V4-Flash & gpt-oss-120b \\
\midrule
accuracy & 0.905 (n=704) & 0.670 (n=763) \\
\quad gold proof uses a universal & 0.773 vs.\ 0.957 & 0.444 vs.\ 0.774 \\
\midrule
claimed length & $-$0.06 [$-$0.92, 0.84] & $-$0.69 [$-$1.43, 0.25] \\
minimum proof size & $-$0.19 [$-$0.96, 0.55] & +0.06 [$-$0.70, 0.64] \\
universal fact used & $\mathbf{-0.79}$ [$-$1.16, $-$0.48] & $\mathbf{-0.47}$ [$-$0.77, $-$0.19] \\
shorter than claimed & $-$0.30 [$-$0.68, 0.09] & $\mathbf{-0.37}$ [$-$0.79, $-$0.04] \\
\midrule
depth measures, jointly & $p=.02$ & $p<.001$ \\
universal fact, beyond depth & $p<.001$ & $p<.001$ \\
\bottomrule
\end{tabular}}
\caption{LLM errors on \fld{}. Top: accuracy on parsed answers (59 DeepSeek replies gave no label and are
excluded; scoring them as errors gives .835). Bottom: standardized logistic
coefficients for correctness, with bootstrap 95\% intervals. \emph{Shorter than claimed}: minimum proof
below the claim without instantiation steps. The depth measures are correlated, so their individual
coefficients are not separately identified; the last two rows are likelihood-ratio tests of the three
depth measures as a block given universal facts, and of universal facts given the three.}
\label{tab:llm-depth}
\end{table}

\paragraph{Computing profiles.} Mandatory steps are decided by one closure and one removal test per step
of a single proof. Min depth, min size and the number of inclusion-minimal proofs are computed by a
provenance chase that tracks a Pareto front of supports and depths. When that front reaches its cap, the
profile is flagged inexact. Min size is then an upper bound and the proof count is approximate; min depth
and mandatory steps stay exact.

\paragraph{\fld{} under its own rules.} \fld{}'s gold proofs use hypothetical syllogism, contraposition
and constructive dilemma as single steps. Our calculus has the first and the last, has modus tollens where
\fld{} has contraposition, and adds disjunctive syllogism and double negation. In \fld{}'s basic rules with
its three derived ones, the picture is the same (0.95 mandatory steps, 41\% of problems with none), and
with contraposition as well even fewer steps are forced (1.37 instead of 2.08 at depth~5).

\paragraph{Boxes as a logic.} The domain is declared with positions at numbered steps and, per
operation type, one rule for what it changes and one for what it leaves alone. Every operation must be
read, so instead of mandatory steps we count the operations whose replacement by ``nothing happens''
changes the answer. The authors' harder split moves a box's contents without naming them.

\paragraph{Audit inputs.} ProofWriter's signed literals and rules over one variable become a ground Horn
program; \fld{} formulas are parsed and certified in the propositional logic's calculus extended with
\fld{}'s simple rules; ZebraLogic grid cells are certified in a grid propagation calculus without case
splits; corrected FOLIO \citep{han2024folio,fixingfolio2026} is parsed into first-order logic and decided
by z3; PlanBench STRIPS instances \citep{valmeekam2023planbench} are searched exactly.

\paragraph{FOLIO.} On the corrected validation split, 112 entailment/contradiction problems from stories
whose premises are consistent parse into first-order logic, and z3 agrees with all their gold labels. Three
stories (12 of the 202 conclusions) are inconsistent as encoded, one only under unique names; of the other 190
conclusions z3 agrees with 186, and the four others need knowledge left unstated, such as ``bees are animals''. FOLIO has no proof level here: its quantifiers, exclusive-or
and biconditionals fall outside our calculi.

\paragraph{ProverQA and LogicGraph.} Both are audited at the premise level from their own formal premises.
Over ProverQA's 1{,}500 development problems z3 agrees with 99\% of labels (1\% are translation errors).
For LogicGraph we enumerate all minimal premise supports of its 900 instances by MARCO on a SAT solver;
enumeration is proved complete on all Small and Medium instances and on 24\% of Large ones within 120\,s
each.

\paragraph{Coverage of proof shapes.} Skeletons are drawn uniformly among the trees of a shape class from
exact counts by height, joins and size. The archive's lead over forward generation (all 140 E/C shape
classes against 78--88) survives size matching. When every mandatory-step count occurs equally often, the
archive covers 125--134 classes against forward's 71--81, and steering rule kinds as well yields more
mandatory rule-kind sets (propositional 157 vs.\ 103; monadic 53 vs.\ 34; modal K 39 vs.\ 29). The same
archive can follow a requested distribution over classes instead of uniform coverage. It then requests the
class furthest below its quota, ignoring failures, and reports classes that never succeed instead of
renormalizing them away: in the grid logic, a quarter to a third of a uniform request is never reached by the constructor (which does not prove those classes impossible), and
the realizable classes are matched within 0.05--0.08 total variation. Premise-level diversity (distinct
required rule-kind multisets) saturates once sources are subsampled to the same needed-premise
distribution (\fld{}, forward and constructed monadic data: 78, 72 and 80 of 85 kinds): sources differ in
depth, not in diversity at a given depth.

\paragraph{Certificate corruption tests.} Countermodels are searched over small domains first, giving
models of 1--14 elements and 3--14\,KB per problem. Over 540 certificates from 9 logics, all pass, and
every injected corruption is caught: altered steps (539/539), deleted mandatory premises (540/540), false
mandatory claims (53/53), omitted mandatory steps (531/531), dropped or swapped countermodels (540/540
each) and altered ones (538/540; the other two altered tables are still models of their query), each
counted over the certificates the corruption applies to. The test covers nine logics (propositional,
monadic, CSP, modal K and S4, nested K and S4, order and kinship); monadic logic with syllogisms, hybrid
modal logic, grids, planning and arithmetic were not included. As an example of the requiredness check,
with contraposition in the rule set, modus tollens is never required.

\begin{table}[t]
\centering\small
\resizebox{\columnwidth}{!}{%
\begin{tabular}{lrrrrr}
\toprule
claimed depth & 1 & 2 & 3 & 4 & 5 \\
\midrule
problems & 128 & 76 & 79 & 24 & 400 \\
claim w/o instantiation & 0.86 & 1.97 & 2.56 & 3.29 & 4.35 \\
min depth & 0.86 & 1.93 & 2.14 & 2.38 & 2.38 \\
min size & 0.86 & 1.93 & 2.56 & 3.17 & 3.08 \\
mandatory & 0.84 & 1.01 & 1.20 & 1.71 & 2.08 \\
minimal proofs & 1.02 & 1.46 & 2.18 & 2.62 & 3.56 \\
\bottomrule
\end{tabular}}
\caption{\fld{} by claimed depth. The smallest proof meets the claim without instantiation steps except
at depth~5; mandatory steps grow far more slowly, and alternative proofs multiply. Depths 1--4 come from a random sample of 400 derivable problems (the others in it: 59 at depth~0, 34 at depths 5--7), depth~5 from a separate sample of 400 depth-5 problems.}
\label{tab:fld-depth}
\end{table}

\paragraph{What a logic can express.} A logic has three layers, and they differ in reach. Its semantics
can be anything a complete solver decides: z3 formulas, a CSP, a Datalog chase. Its calculus is a set of
monotone rule schemas over typed terms, with unification and symmetry handled once for all logics; this
covers forward reasoning in every logic listed, and adding one is a matter of declaring symbols and rules
(hybrid modal logic and state tracking over boxes were added without any change to the machinery;
for state tracking, where every event must be read, what a question requires is the set of events whose
replacement by an idle step loses the answer). The calculus has no scoped
assumptions (conditional proof, reductio, case splits), no computed terms such as $a+b$, and no side
conditions beyond typing, which is why hypothetical \fld{} problems and some ZebraLogic cells are out of
reach. Construction, finally, depends on how redundant the calculus is: shapes are built reliably where
shortcuts are few and recurrent, and fail where a quantity can be derived many ways, as in arithmetic
over scores. Scoped subproofs are the extension that would widen the second layer most, at the price of a
mandatory-step analysis over assumptions.

\begin{table}[t]
\centering\small
\resizebox{\columnwidth}{!}{%
\begin{tabular}{lrrrrrr}
\toprule
& & forward & filter & \multicolumn{3}{c}{success (seconds per certified instance)} \\
\cmidrule(lr){5-7}
logic & $n$ & all needed & s/instance & reject, 8 & reject, 64 & ours \\
\midrule
prop. & 8 & 52\% & 0.010 & 100\% (0.013) & -- & 100\% (0.010) \\
prop. & 30 & 5\% & 0.27 & 56\% (0.40) & 98\% (0.22) & 100\% (0.058) \\
monadic & 8 & 50\% & 0.020 & 99\% (0.021) & -- & 99.5\% (0.021) \\
monadic & 30 & 15\% & 0.26 & 67\% (0.29) & 99\% (0.17) & 99\% (0.081) \\
CSP & 8--30 & 100\% & 0.02--0.04 & 100\% & -- & 100\% \\
\bottomrule
\end{tabular}}
\caption{Building instances with $n$ applications, all certified mandatory. \emph{Forward all needed}:
share of forward-generated instances whose planned steps are all mandatory; \emph{filter}: forward
generation kept only when all steps are mandatory; \emph{reject}: backward construction with
per-instance rejection and no atlas, with at most 8 or 64 redraws per request (ours: 8). 200 instances or
requests per line (CSP: 20; 64 redraws: 100, where the atlas takes 0.064 and 0.086\,s). CSP propagation
has no shortcuts and serves as the control.}
\label{tab:baselines}
\end{table}

\paragraph{Several proofs requested.} When $k\ge2$ proofs are requested, only the requested bottlenecks must be mandatory, since a fully
mandatory plan is the unique minimal proof (line~\ref{alg:accept} of Algorithm~\ref{alg:proofplant}).

\section{Encoder Experiments}
\label{app:encoders}

\paragraph{Protocol.} Each run fine-tunes a pretrained encoder with a shared body and one classification
head per task, on a procedural source and the target's 200 training examples jointly, for 600 optimizer
steps (batch 4, gradient accumulation 4, learning rate $3\cdot10^{-5}$, bf16), with batches drawn from
source and target in equal numbers. The baseline trains on the target examples alone for the same number of
steps. Every training set shares the baseline of its seed, so training sets compare as paired per-seed differences. Targets
are \fld{}'s held-out test, semantic fragments \citep{richardson2020fragments}, WANLI
\citep{liu2022wanli}, FOLIO and held-out constructed examples; encoders are ModernBERT-base
\citep{warner2024modernbert} and DeBERTa-v3-base \citep{he2023debertav3}.

\paragraph{Constructed vs.\ \fld{} at equal size.}
With ModernBERT and three seeds, constructed monadic data beats \fld{}'s own training data on \fld{}'s
held-out test (45.5$\pm$0.5 vs.\ 44.0$\pm$1.7, 3/3 seeds) and on semantic fragments (50.7 vs.\ 45.6), and
ties on WANLI (47.6 vs.\ 48.1). Over five seeds, the paired per-seed differences, constructed minus \fld{},
are +2.2$\pm$1.3 on \fld{} held-out and +6.5$\pm$2.0 on semantic fragments (5/5 seeds each), +3.9$\pm$4.1
on FOLIO (4/5) and $-$1.0$\pm$2.7 on WANLI (2/5). Training on \fld{} itself barely moves its own held-out
test (+0.2 over the baseline). With DeBERTa-v3 the \fld{} held-out lead replicates (+1.1$\pm$1.6, 5/5);
the other targets are noisier (Table~\ref{tab:necessity}); single DeBERTa-v3 seeds collapse at this
budget (60\% in-domain for one constructed seed). Shape-constructed and forward-generated examples matched
on certified structure do not separate outside the domain (all differences within one standard
deviation), as with LLMs (Appendix~\ref{app:llm}).

\paragraph{Paired necessity training sets.} Adding to each certified example one premise that either
bypasses most of its proof or, with constants renamed, leaves it intact gives training sets that differ in
mandatory steps (1.77 vs.\ 10.82) and in smallest proof size at matched planted depth, length and label;
they tie within a standard deviation except on \fld{} held-out, where the two encoders disagree in sign
(Table~\ref{tab:necessity}). Constructed monadic data beats both on semantic fragments (+7.7) and in-domain
(+33.4 against +20), so the gap points to the distribution of rule kinds and shapes, not to depth.

\begin{table}[t]
\centering\small
\resizebox{\columnwidth}{!}{%
\begin{tabular}{lcccc}
\toprule
& \multicolumn{2}{c}{necessary $-$ bypass} & \multicolumn{2}{c}{constructed $-$ \fld{}} \\
\cmidrule(lr){2-3}\cmidrule(lr){4-5}
target & ModernBERT & DeBERTa-v3 & ModernBERT & DeBERTa-v3 \\
\midrule
\fld{} held-out & +1.1$\pm$1.1 (4) & $-$1.8$\pm$1.9 (1) & +2.2$\pm$1.3 (5) & +1.1$\pm$1.6 (5) \\
semantic fragments & $-$0.5$\pm$3.1 (2) & +0.5$\pm$7.0 (2) & +6.5$\pm$2.0 (5) & +6.1$\pm$10.3 (3) \\
WANLI & $-$0.8$\pm$0.6 (0) & $-$0.5$\pm$1.9 (3) & $-$1.0$\pm$2.7 (2) & $-$2.2$\pm$2.5 (1) \\
FOLIO (corrected) & +0.2$\pm$2.2 (1) & $-$0.8$\pm$3.1 (2) & +3.9$\pm$4.1 (4) & $-$1.8$\pm$3.7 (2) \\
constructed held-out & $-$0.3$\pm$3.2 (3) & $-$0.7$\pm$5.5 (2) & +31.9$\pm$6.7 (5) & +15.4$\pm$22.1 (4) \\
\bottomrule
\end{tabular}}
\caption{Encoders: paired per-seed differences in accuracy (points; mean $\pm$ sample standard deviation
over 5 seeds; in parentheses, seeds where the difference is positive). The necessity training sets differ in
mandatory steps (10.82 vs.\ 1.77) and, through the bypass premise, in smallest proof size.}
\label{tab:necessity}
\end{table}

\section{LLM Experiments}
\label{app:llm}

\begin{table}[t]
\centering\small
\resizebox{\columnwidth}{!}{%
\begin{tabular}{lcccccc}
\toprule
& \multicolumn{3}{c}{FOLIO} & ModalBench & \fld{} & counterf. \\
\cmidrule(lr){2-4}
training sources & macro & N recall & N$\to$C & E/C & macro & all three \\
\midrule
monadic (3k examples) & .46 & .24 & .25 & .53 & .62 & .18 \\
+ modal + Kripke (3k) & .48 & .12 & .33 & .61 & .62 & .22 \\
+ modal + Kripke (9k, selected) & .50 & .14 & .38 & \textbf{.65} & .62 & -- \\
+ modal + partial Kripke (9k) & \textbf{.51} & \textbf{.32} & .26 & .64 & \textbf{.63} & \textbf{.34} \\
\bottomrule
\end{tabular}}
\caption{Qwen3-0.6B on certified sources, seeds 10--12. ModalBench: held-out two-thirds, neutral
answers replaced by the likelier of E and C. N recall / N$\to$C: FOLIO neutrals answered neutral /
contradiction. Counterfactual: share of 201 groups with all three examples right.}
\label{tab:sources}
\end{table}

\begin{table*}[t]
\centering\small
\setlength{\tabcolsep}{5pt}
\begin{tabular}{lccccccccc}
\toprule
training data & FOLIO & ModalB. & \fld{} & ProverQA & ProofW.$^\ast$ & TPTP & QMLTP & W-A & N gap \\
\midrule
\multicolumn{10}{l}{\emph{Qwen3-0.6B, 3{,}000 examples}} \\
certified types, searched & \textbf{.68} & .62 & .65 & .62 & .86 & .78 & .74 & \emph{.69} & .14 \\
certified types, uniform & .65 & .61 & .63 & .60 & .78 & .72 & .69 & \emph{.72} & .13 \\
planted-depth types, searched & \textbf{.68} & .58 & .63 & .60 & .88 & .78 & .71 & \emph{.70} & \textbf{.12} \\
planted-depth types, uniform & .67 & \textbf{.63} & .63 & .60 & .81 & .69 & .71 & \emph{.77} & .14 \\
\fld{}-style, same logics & .62 & .58 & .59 & .55 & \textbf{.94} & .77 & .58 & .61 & .21 \\
\cmidrule(lr){1-10}
\fld{} & .61 & .53 & \emph{.82} & \textbf{.64} & .91 & \textbf{.84} & .62 & .59 & .36 \\
ProverQA & .63 & .54 & .62 & \emph{.80} & .93 & .64 & .49 & .57 & .36 \\
ProofWriter & .58 & .57 & \textbf{.66} & .60 & \emph{.99} & \textbf{.84} & \textbf{.75} & .57 & .28 \\
\midrule
\multicolumn{10}{l}{\emph{Qwen3-0.6B, 9{,}000 examples}} \\
certified types, searched & \textbf{.73} & .65 & \textbf{.63} & .62 & .90 & \textbf{.76} & \textbf{.68} & \emph{.74} & .11 \\
certified types, uniform & .71 & \textbf{.66} & \textbf{.63} & \textbf{.63} & .81 & .68 & \textbf{.68} & \emph{.80} & .11 \\
\fld{}-style, same logics & .66 & .59 & .58 & .57 & \textbf{.94} & .72 & .64 & .61 & .11 \\
\midrule
\multicolumn{10}{l}{\emph{Qwen3-1.7B, 3{,}000 examples}} \\
certified types, searched & \textbf{.73} & \textbf{.64} & \textbf{.70} & \textbf{.71} & .85 & .79 & .77 & \emph{.73} & \textbf{.14} \\
certified types, uniform & .70 & \textbf{.64} & .69 & \textbf{.71} & .83 & .75 & .71 & \emph{.75} & .15 \\
\fld{}-style, same logics & .71 & .59 & .65 & .68 & \textbf{.86} & \textbf{.82} & \textbf{.81} & .65 & .18 \\
\midrule
\multicolumn{10}{l}{\emph{Qwen3-4B, 3{,}000 examples}} \\
certified types, searched & \textbf{.84} & .74 & \textbf{.73} & .79 & .77 & \textbf{.86} & \textbf{.91} & \emph{.79} & .13 \\
certified types, uniform & \textbf{.84} & \textbf{.75} & .72 & \textbf{.80} & .74 & .81 & .87 & \emph{.82} & \textbf{.12} \\
\fld{}-style, same logics & .82 & .68 & .68 & .74 & \textbf{.89} & .82 & .81 & .66 & .14 \\
\midrule
\multicolumn{10}{l}{\emph{Qwen3-8B, 3{,}000 examples}} \\
certified types, searched & .86 & .74 & \textbf{.75} & .78 & .83 & \textbf{.89} & \textbf{.92} & \emph{.82} & .11 \\
certified types, uniform & \textbf{.87} & \textbf{.75} & .74 & \textbf{.79} & .78 & .86 & .91 & \emph{.82} & .11 \\
\fld{}-style, same logics & .85 & .69 & .72 & .77 & \textbf{.87} & .86 & .90 & .66 & .11 \\
\midrule
\multicolumn{10}{l}{\emph{SmolLM2-1.7B, 3{,}000 examples}} \\
certified types, searched & .66 & .50 & .63 & .63 & .90 & \textbf{.82} & \textbf{.54} & \emph{.67} & \textbf{.14} \\
certified types, uniform & \textbf{.67} & .50 & \textbf{.66} & \textbf{.67} & .89 & .70 & .51 & \emph{.65} & .18 \\
\fld{}-style, same logics & .66 & \textbf{.51} & \textbf{.66} & .65 & \textbf{.93} & .62 & .51 & .63 & .36 \\
\bottomrule
\end{tabular}
\caption{Two-way accuracy (E vs.\ C) on untouched tests at equal training size. Qwen3-0.6B (full fine-tuning) and
Qwen3-1.7B, 4B, 8B and SmolLM2-1.7B (LoRA, rank 16); mean of seeds 10--16. The mixture was
searched once (0.6B, 3{,}000 examples) and reused unchanged. Planted-depth types: the same nine types built from forward examples binned by planted depth instead of mandatory steps. Italics: test from the training source; bold: best
of the others in its block. W-A: world-assumption pairs, Kripke pairs differing only in the sentence that closes the world. \fld{}-style examples are derivations only; about a quarter of the certified mixture is Kripke model checking.
N gap: mean absolute gap between the share of neutral answers and the share of neutral examples, over the
eight tests. $^\ast$ProofWriter's labels follow hypothesis negation (the cue alone scores .945), so its
column partly measures reliance on that cue (Appendix~\ref{app:polarity}).}
\label{tab:search}
\end{table*}

\paragraph{Shortcuts the controls removed.}\label{app:shortcuts} Each naive comparison we ran was first decided by a shortcut
that a matched control then removed. When deep and shallow training sets shared their neutral examples, depth alone told the model which
answers were neutral, and the depth result reversed (deep 0.78 vs.\ shallow 0.45 at 13+ steps); pairing
forward-generated E/C examples with our constructed neutrals let the generator's style give the label away (0.32 on \fld{}, 0.60 with forward-style
neutrals). A hypothesis-only TF-IDF probe reached .48 (chance .33) because our monadic calculus could not
refute a universal; every training set shares the cue, and two refutation rules bring it to .39 on fresh examples.
The metric is a shortcut too: \fld{}'s test and ModalBench have no neutral problems, so macro accuracy over
three labels rewards a training set that merely makes the model answer neutral less. A mixture search
that maximized it found such a mixture (neutral answers on 2\% of \fld{} examples against 12\% for uniform
weights), which then failed the world-assumption test below (.44 against .68).

\begin{figure}[t]
\centering
\includegraphics[width=\columnwidth]{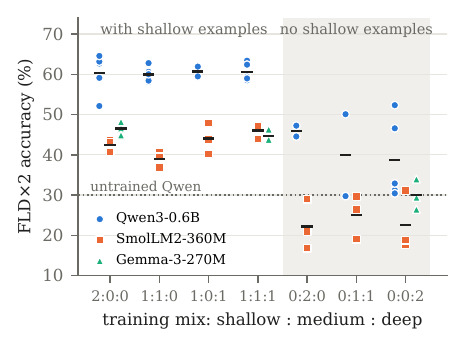}
\caption{Training with and without shallow examples, scored on \fld{} (accuracy on its entailment and
contradiction problems, averaged over the two labels; a neutral answer counts as wrong). Each training set
mixes shallow (1--3 mandatory steps), medium (4--7) and deep (8+) examples in the proportions shown: 2:0:0
is shallow only, 1:1:1 equal thirds. One dot per seed (Qwen: seeds 0--1 and 10--12; SmolLM2, Gemma:
10--12), bars at the mean. Every mixture with shallow examples beats every mixture without them on SmolLM2
and Gemma; on Qwen, mixtures without them vary widely across seeds.}
\label{fig:mix}
\end{figure}

\paragraph{Depth mixtures: protocol.} We fine-tune Qwen3-0.6B-Base, SmolLM2-360M and Gemma-3-270M \citep{gemma2025gemma3} with full
fine-tuning (2 epochs, 3{,}000 monadic examples), scoring labels by log-probability (macro accuracy). Weights over the three
depth ranges are chosen on validation sets and confirmed on fresh seeds (10--12) and untouched tests
(Table~\ref{tab:llm-mix}, per seed in Figure~\ref{fig:mix}). Finer rankings than shallow vs.\ deep-only do
not transfer between models (Spearman 0.41), and SmolLM2 and Gemma stay near chance on FOLIO.

\begin{table}[t]
\centering\small
\resizebox{\columnwidth}{!}{%
\begin{tabular}{lccccccc}
\toprule
& \multicolumn{5}{c}{Qwen3-0.6B} & SmolLM2 & Gemma-3 \\
\cmidrule(lr){2-6}
training examples (shallow:mid:deep) & in-logic & FOLIO & \fld{} & N & commit & \fld{} & \fld{} \\
\midrule
constructed 2:0:0 & \textbf{.77} & \textbf{.47} & \textbf{.62} & .02 & .64 & .43 & \textbf{.47} \\
constructed 1:1:0 & \textbf{.77} & .46 & .61 & .05 & .64 & .39 & -- \\
constructed 1:1:1 & .76 & .45 & .60 & .06 & .64 & \textbf{.46} & .45 \\
constructed 0:0:2 & .72 & .40 & .32 & .53 & .68 & .23 & .30 \\
\midrule
forward + filter 1:1:1 & .73 & -- & .32 & .49 & .64 & -- & -- \\
\quad with forward-style neutrals & .74 & -- & .60 & .10 & .66 & .17 & .32 \\
\bottomrule
\end{tabular}}
\caption{LLMs trained on mixtures of mandatory-step ranges: macro accuracy, mean of seeds 10--12. On \fld{}, N is the
share of derivable problems answered neutral and \emph{commit} the accuracy of the answers that are E or C:
deep-only training loses by abstaining, not by confusing the two labels (\fld{} ranges
$\pm$.03 for constructed training sets on Qwen). \emph{Forward + filter}: \fld{}-style forward examples certified and
filtered into the same depth ranges. FOLIO: corrected validation set (202 examples). Untrained Qwen: .34 in-logic,
.32 FOLIO, .30 \fld{}. Qwen deep-only is seed-sensitive (five-seed mean .39 on \fld{}).}
\label{tab:llm-mix}
\end{table}

\paragraph{Construction vs.\ filtered forward examples.} With matched neutrals, forward examples filtered to the
same depth ranges equal constructed ones on Qwen (0.60 vs.\ 0.60) but trail on the two smaller models, which
call 42--73\% of derivable \fld{} examples neutral. Forced to choose entailment or contradiction, SmolLM2's
forward training set matches the constructed one (0.53 vs.\ 0.54): the gap is the neutral decision, and both
smaller models are near chance on \fld{}'s direction, so only Qwen's transfer measures reasoning.

\paragraph{Writing proofs.} Models can also be trained to write proofs instead of labels, and then graded
on the proofs. Premises are numbered; the target is a proof whose steps cite the premises they use,
followed by the answer. A proof is \emph{grounded} when its cited premises alone give the answer (by the
logic's semantics; for \fld{} and ProverQA by z3 over their formulas, for ProofWriter by chasing its
rules) and \emph{minimal} when none of them can be dropped. The intermediate steps are not checked, so
grounding measures premise selection, not the validity of each inference.
Training sets have 4{,}500 examples (3{,}000 entailment and contradiction, 1{,}500 neutral; seeds 10--12)
and are otherwise matched as in Section~\ref{sec:transfer}. Certified problems with their smallest proofs give models whose proofs of
held-out certified problems are grounded twice as often as models trained on forward problems in the same
format (Table~\ref{tab:proofs}), and the gap is concentrated among deeper problems: on problems with at most five mandatory steps
every training set does equally well, while past ten mandatory steps the forward-trained models rarely
produce a proof that holds. The gap is not citation volume: citing a random set of premises of the same
size grounds .29 (certified) and .12 (forward) of answers, and the certified lead holds for minimal
citations (.52 against .29), which citing every premise cannot obtain. Replacing the smallest proof by the planted one on the same forward problems
changes little; in this logic the two have the same size. Outside the training logic grounded proofs are
rare for every training set at this model size, and only \fld{}'s own proofs do well on \fld{}. A larger
model trained on several logics at once (Qwen3-1.7B with LoRA, the same 3{,}000 problems split over
propositional, monadic, order, kinship and modal logic; seeds 10--12) roughly doubles transfer on shallow
problems, for certified and forward training alike: grounded proofs of ProverQA's easy problems rise from
.25 to .46 (certified) and of depth-0 \fld{} problems from .19 to .42. Proofs of deeper benchmark problems
stay rare (at most .08 on ProofWriter beyond depth~1 and on ProverQA's hard problems), and with the
training problems split across logics the certified lead on held-out monadic problems narrows (.27
against .21). Writing a
proof also costs label accuracy against training on answers alone, as generating a proof before the label
did above, and the certified model tends to answer neutral problems with a proof. With labels as the answer,
training on the certified proof after the label leaves accuracy unchanged, and generating it before the
label gains .03 in the logic but loses .07 on \fld{}, whose proofs end in the hypothesis as stated as often
for contradictions as for entailments.

\begin{table}[t]
\centering\small
\resizebox{\columnwidth}{!}{%
\begin{tabular}{lccccccc}
\toprule
& \multicolumn{3}{c}{held-out monadic} & & & & \\
\cmidrule(lr){2-4}
training proofs & all & 11--15 & 16+ & propos. & \fld{} & ProofW. & ProverQA \\
\midrule
certified problems, smallest proof & \textbf{.63} & \textbf{.67} & \textbf{.24} & .05 & .17 & .10 & .12 \\
forward problems, smallest proof & .30 & .12 & .02 & .05 & .20 & .11 & .12 \\
forward problems, planted proof & .37 & .27 & .05 & .05 & .20 & .13 & \textbf{.15} \\
\fld{} proofs & .02 & .01 & .00 & .02 & \emph{.57} & \textbf{.17} & .07 \\
\bottomrule
\end{tabular}}
\caption{Grounded proofs: share of entailment and contradiction test problems answered correctly with a
proof whose cited premises alone give the answer, checked exactly. Qwen3-0.6B, mean of seeds 10--12;
11--15 and 16+: certified mandatory steps. Minimal proofs follow the same pattern on the monadic tests
(.53, .29, .36, .02); on \fld{} and ProofWriter almost no proof from constructed training is minimal,
because the cited sets include distractors. Italics: test from the training source.}
\label{tab:proofs}
\end{table}

\begin{figure}[t]
\centering
\includegraphics[width=\columnwidth]{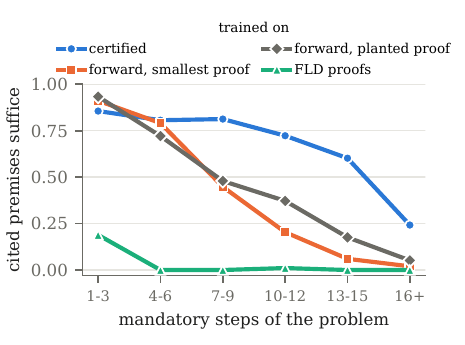}
\caption{Share of problems answered correctly with a proof whose cited premises alone give the answer
(intermediate steps unchecked), on held-out certified monadic problems, by the problem's mandatory steps. Qwen3-0.6B, 4{,}500 training examples in the
same proof format, mean of three seeds (Table~\ref{tab:proofs}).}
\label{fig:proofs}
\end{figure}

\paragraph{Sources: reading assumptions and counterfactuals.} On 300 Kripke pairs that differ only in the
sentence closing the world, the mixture with partially observed Kripke examples changes its answer exactly
when it should in four pairs of five, the others at chance. One failure is shared by every mixture: models
keep answering entailment when a mandatory premise is removed, also under paraphrase, so none gets a full
counterfactual group (entailment, missing premise, negated hypothesis) right more than a third of the
time.

\paragraph{Mixture search: protocol.} The 60 proxies are trained on random 3{,}000-example mixtures of the
nine problem types. The regression target is the mean score over the development splits of \fld{}, FOLIO,
ModalBench, ProverQA and ProofWriter, scored with the label prior factored out. Development scores span
.48--.64, and leave-one-out predictions correlate .55 with them. Tasks agree on what helps, except
ProofWriter, which barely correlates with the others. It beats the best of the 60
proxies mainly on tests no selection saw. Carried to other model sizes and to 9{,}000 examples, it keeps its gains on ProofWriter
and the prover-derived tests and is otherwise level with uniform weights. The nine problem types are
monadic and labelled modal examples in three ranges of mandatory steps (1--3, 4--7 and 8+), closed Kripke examples of modal depth 0--1 and 2--3 (E/C only), and partially observed Kripke
examples (modal depth 0--3, each fact hidden with probability 0.3, E/C/N in thirds). The full selected
weights, in that order, are .292, .289, .003; .037, .011, .119; .001, .109; .140. The regression is a ridge
regression of the mean development score on the nine weights; with the label prior factored out, a development set
without neutral problems is scored by the likelier of E and C, and one with neutral problems by macro
accuracy under the best pair of per-label biases fitted on that set, so that a mixture is not rewarded for
a label prior that happens to suit it.

\paragraph{Polarity cue in the external tests.}\label{app:polarity} A test is cue-heavy when one surface feature
nearly gives the label. For each test, we count how often the E/C problems with a negated hypothesis (a
negation word in the hypothesis sentence) are contradictions. Predicting contradiction exactly when the
hypothesis is negated scores (two-way) .945 on ProofWriter, .864 on TPTP, .773 on QMLTP, and .50--.60 on
every other test. Training sets carry the cue to different degrees: the gap between the negation rates of
contradictions and entailments is +.90 in ProofWriter's training data, +.28 in the \fld{}-style set, +.16 in
\fld{}'s and +.13 in the certified mixture. On the three cue-heavy tests, Table~\ref{tab:polarity} splits
the certified-minus-\fld{}-style difference into problems where the cue gives the right answer and
problems where it gives the wrong one. \fld{}-style examples win on the first kind and lose on the second at
most sizes. On QMLTP, all contradictions are negated, so the cue misleads only on entailments.

The training skew does not cause the \fld{}-style pattern. We rebuilt both sets with the same problem types
and the same label counts per type, choosing within each type and label half the examples with a negated
hypothesis (0.6B, seeds 10--16). The balanced \fld{}-style set keeps its ProofWriter score (.930 against .935) and matches the original on cue-right and cue-wrong
problems of every test, except small gains where the cue misleads on FOLIO and ModalBench (+.07, +.06).
The balanced certified mixture still leads it where the cue misleads: by .20 on ProofWriter, .14 on TPTP,
and on the tests the cue barely predicts, by .21 on ModalBench, .19 on ProverQA and .09 on FOLIO. Balancing the certified mixture itself trades
cue-right for cue-wrong problems (ProofWriter $-$.04 and +.07, ModalBench $-$.12 and +.08) and lowers its
overall ProofWriter score (.83 against .86), so we keep the published mixture.

The certified-depth contrast of \S\ref{sec:transfer} carries the cue unevenly: the negation gap is .34 in
the certified-deep set and .13 in the shallow one, because deep proofs more often end in a negated
contradiction. Deep minus shallow, on problems where the cue is right and where it misleads (fourteen seeds,
bold as in Table~\ref{tab:polarity}): QMLTP \textbf{+.19} and \textbf{$-$.15}, TPTP +.05 and \textbf{$-$.14},
\fld{} \textbf{+.09} and \textbf{$-$.05}, FOLIO \textbf{+.03} and \textbf{$-$.06}, ProverQA \textbf{+.02} and
\textbf{$-$.05}, world-assumption pairs +.05 and \textbf{$-$.07}, ProofWriter +.01 and $-$.04, ModalBench
+.04 and +.02. Rebuilt with half the examples of each label and planted depth negated in both sets (same
generator, same matching), the asymmetry disappears: the only interval that excludes zero is ProofWriter's 22
cue-wrong problems (+.08), and QMLTP keeps a trend (+.08 and $-$.06, both intervals touching zero).
Table~\ref{tab:ablations} reports the balanced contrast.

\begin{table}[t]
\centering\small
\setlength{\tabcolsep}{3pt}
\begin{tabular}{llcccc}
\toprule
test & rows & 0.6B & 1.7B & 4B & 8B \\
\midrule
ProofWriter & cue right (378) & \textbf{$-$.08} & $-$.01 & \textbf{$-$.12} & \textbf{$-$.05} \\
 & cue wrong (22) & \textbf{+.17} & $-$.03 & +.05 & \textbf{+.10} \\
TPTP & cue right (890) & .00 & $-$.02 & \textbf{+.03} & \textbf{+.02} \\
 & cue wrong (140) & \textbf{+.08} & $-$.06 & \textbf{+.09} & \textbf{+.06} \\
QMLTP & cue right (464) & \textbf{+.25} & \textbf{$-$.06} & \textbf{+.12} & +.01 \\
 & cue wrong (136) & $-$.16 & .00 & +.03 & \textbf{+.07} \\
\bottomrule
\end{tabular}
\caption{Certified mixture minus \fld{}-style examples (two-way accuracy, Qwen3, seeds 10--16), on problems
where the polarity cue gives the right label and where it gives the wrong one. Bold: 95\% interval
resampling seeds and test problems excludes zero.}
\label{tab:polarity}
\end{table}

\paragraph{Controlled contrasts in full.} Table~\ref{tab:ablations} gives every test for the three
contrasts that isolate a structural variable: mandatory steps at fixed planted depth, one proof against
several at fixed proof length (forward-generated monadic problems with exactly one or several minimal
proofs, matched per label and shortest proof at 6.6 steps, so that the many-proof problems
force 3.7 steps against 6.6), and certified against forward derivation examples at fixed mixture
composition (the forward examples of the same pack and label replace the derivation examples of the
searched mixture as the generator makes them, with no binning by depth).

\begin{table*}[t]
\centering\small
\setlength{\tabcolsep}{4pt}
\begin{tabular}{lcccccc}
\toprule
& \multicolumn{2}{c}{many $-$ few mandatory steps, balanced (14)} & \multicolumn{2}{c}{several proofs $-$ one (14)} & \multicolumn{2}{c}{certified $-$ forward derivations (7)} \\
\cmidrule(lr){2-3}\cmidrule(lr){4-5}\cmidrule(lr){6-7}
test & two-way & N rate & two-way & N rate & two-way & N rate \\
\midrule
FOLIO & $-$.012 [$-$.047, .021] & .48 $\to$ .66 & +.008 [$-$.023, .040] & .28 $\to$ .26 & +.022 [$-$.015, .061] & .10 $\to$ .09 \\
ModalBench & +.003 [$-$.013, .020] & .39 $\to$ .30 & $-$.002 [$-$.022, .019] & .27 $\to$ .30 & +.015 [$-$.009, .042] & .00 $\to$ .01 \\
\fld{} & +.011 [$-$.007, .030] & .34 $\to$ .34 & \textbf{$-$.021} [$-$.040, $-$.003] & .17 $\to$ .21 & +.014 [$-$.007, .034] & .04 $\to$ .02 \\
ProverQA & $-$.007 [$-$.026, .009] & .48 $\to$ .58 & $-$.011 [$-$.030, .006] & .37 $\to$ .35 & .000 [$-$.017, .019] & .27 $\to$ .28 \\
ProofWriter & $-$.004 [$-$.028, .018] & .07 $\to$ .14 & \textbf{+.027} [.007, .049] & .12 $\to$ .04 & +.007 [$-$.019, .031] & .03 $\to$ .02 \\
TPTP & +.025 [$-$.019, .070] & .62 $\to$ .64 & $-$.024 [$-$.063, .012] & .55 $\to$ .50 & +.005 [$-$.019, .028] & .19 $\to$ .14 \\
QMLTP & +.049 [$-$.003, .103] & .63 $\to$ .63 & $-$.017 [$-$.076, .039] & .35 $\to$ .46 & +.039 [$-$.020, .102] & .01 $\to$ .01 \\
W-A pairs & $-$.016 [$-$.038, .006] & .46 $\to$ .49 & +.007 [$-$.013, .025] & .32 $\to$ .34 & $-$.016 [$-$.047, .016] & .06 $\to$ .06 \\
\bottomrule
\end{tabular}
\caption{The three controlled contrasts on every test (Qwen3-0.6B). The depth contrast has negated hypotheses balanced per label and planted depth in both sets (the unbalanced version and its polarity confound are in Appendix~\ref{app:polarity}). Two-way accuracy difference with 95\%
intervals that resample seeds and test problems (bold: excludes zero), and the share of neutral answers
from the first training set to the second. \fld{} and ModalBench have no neutral problems; on the other
tests a third or fewer are neutral.}
\label{tab:ablations}
\end{table*}

\end{document}